\documentclass[a4paper,fleqn]{cas-dc}

\usepackage[authoryear]{natbib}
\usepackage{verbatim}
\usepackage{graphicx}
\usepackage{bm}
\usepackage{booktabs}
\usepackage{multirow}
\usepackage{tabularx}

\usepackage{hyperref}
\usepackage{cleveref}

\usepackage{algorithm}
\usepackage{algpseudocode}
\usepackage{amsmath}
\usepackage{color}

\def\tsc#1{\csdef{#1}{\textsc{\lowercase{#1}}\xspace}}
\tsc{WGM}
\tsc{QE}

\begin{document}
\begin{sloppypar}
\let\WriteBookmarks\relax
\def\floatpagepagefraction{1}
\def\textpagefraction{.001}

\shorttitle{OGMN: Occlusion-guided Multi-task Network for Object Detection in UAV Images} 

\shortauthors{Xuexue Li, et al}  

\title [mode = title]{OGMN: Occlusion-guided Multi-task Network for Object Detection in UAV Images} 



\author[1,2,3,4]{Xuexue Li}
\ead[ORCID]{0000-0002-0177-7001}
\ead{lixuexue20@mails.ucas.ac.cn}

\author[1,2,3,4]{Wenhui Diao}
\cormark[1]
\ead{diaowh@aircas.ac.cn}

\author[1,2,3,4]{Yongqiang Mao}

\author[5]{Peng Gao}

\author[5]{Xiuhua Mao}

\author[1,2,3,4]{Xinming Li}

\author[1,2,3,4]{Xian Sun}

\address[1]{Aerospace Information Research Institute, Chinese Academy of Sciences, Beijing 100190, China}


\address[2]{Key Laboratory of Network Information System Technology (NIST), Aerospace Information Research Institute, Chinese Academy of Sciences, Beijing 100190, China }

\address[3]{University of Chinese Academy of Sciences, Beijing 100190, China}

\address[4]{School of Electronic, Electrical and Communication Engineering, University of Chinese Academy of Sciences, Beijing 100049, China}

\address[5]{Beijing Institute of Tracking and Telecommunication Technology, China}
\cortext[1]{Corresponding author}

\begin{abstract}
Occlusion between objects is one of the overlooked challenges for object detection in UAV images. Due to the variable altitude and angle of UAVs, occlusion in UAV images happens more frequently than that in natural scenes. Compared to occlusion in natural scene images, occlusion in UAV images happens with feature confusion problem and local aggregation characteristic. And we found that extracting or localizing occlusion between objects is beneficial for the detector to address this challenge. According to this finding, the occlusion localization task is introduced, which together with the object detection task constitutes our occlusion-guided multi-task network (OGMN). 
The OGMN contains the localization of occlusion and two occlusion-guided multi-task interactions. In detail, an occlusion estimation module (OEM) is proposed to precisely localize occlusion. Then the OGMN utilizes the occlusion localization results to implement occlusion-guided detection with two multi-task interactions. One interaction for the guide is between two task decoders to address the feature confusion problem, and an occlusion decoupling head (ODH) is proposed to replace the general detection head. Another interaction for guide is designed in the detection process according to local aggregation characteristic, and a two-phase progressive refinement process (TPP) is proposed to optimize the detection process.
Extensive experiments demonstrate the effectiveness of our OGMN on the Visdrone and UAVDT datasets. In particular, our OGMN achieves 35.0\% mAP on the Visdrone dataset and outperforms the baseline by 5.3\%. And our OGMN provides a new insight for accurate occlusion localization and achieves competitive detection performance.
\end{abstract}

\begin{keywords}
\sep Object detection
\sep UAV image
\sep Multi-task learning
\sep Occlusion localization
\sep Multi-task interaction
\end{keywords}

\maketitle

\section{Introduction}
Object detection is an important task in computer vision, which is widely used in frontier fields such as visual-and-language navigation \citep{wu2021visual}, video captioning \citep{li2019visual}, video representation learning \citep{fujitake2022video}, and visual tracking \citep{smeulders2013visual}, etc. And object detection based on deep learning architecture makes a great breakthrough, which is the mainstream pipeline currently. For natural scenes such as MS COCO \citep{lin2014microsoft} and PASCAL VOC \citep{everingham2010pascal}, general detectors \citep{law2018cornernet,ge2021yolox,cai2018cascade} perform well in natural scenes, but when migrating to UAV images, they suffer from scale variation, uneven distribution and occlusion problems in UAV images. Although recent studies \citep{ozge2019power,wang2019spatial,zhang2019fully,li2020density,yang2019clustered,zhou2019scale,deng2020global} on UAV object detection improve model robustness on scale variation and uneven distribution problems, they overlooked occlusion problem involved, resulting in still-limited detection performance. 

\begin{figure}[b]
\centering
\includegraphics[scale=0.325]{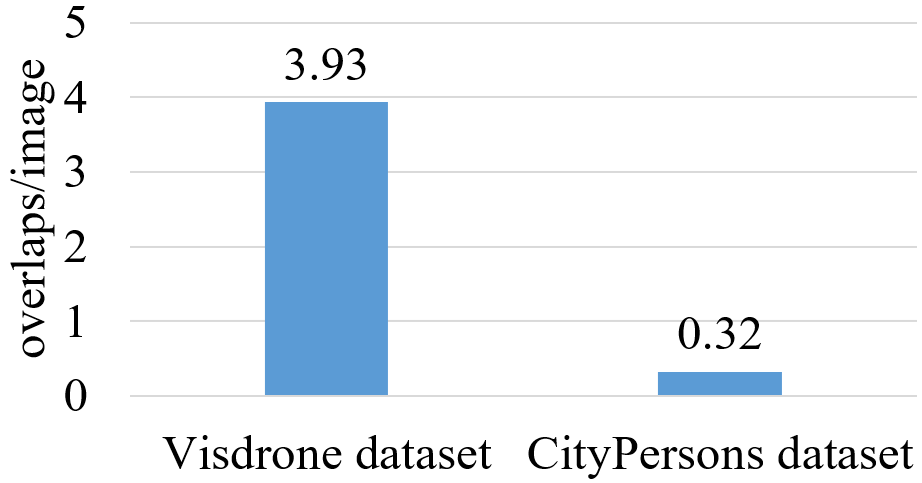}
\caption{The comparison of the occlusion frequency among natural scene and UAV scene. Each overlap indicates that occlusion happens between two objects. The occlusion in UAV scenes such as the Vidrone dataset happens about ten times more than occlusion in the natural scene such as the CityPersons dataset.}
\label{figa}
\end{figure}

\begin{figure}[t]
\centering
\includegraphics[scale=0.702]{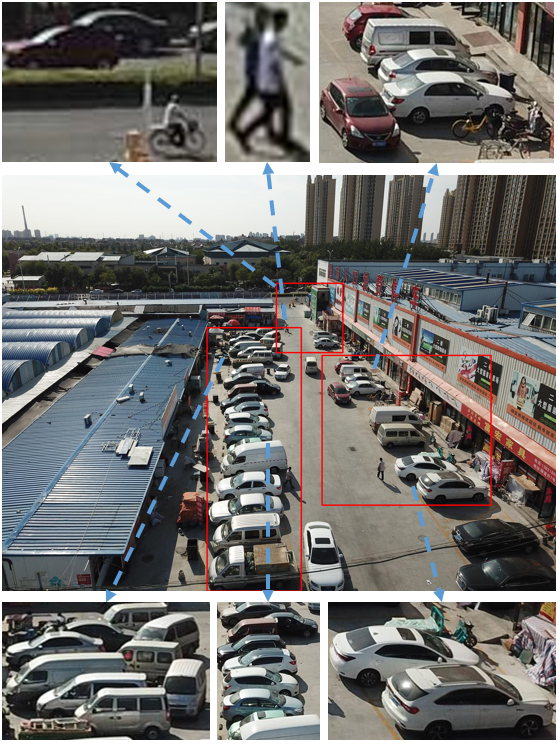}
\caption{The occlusion in UAV images happens frequently with two characteristics, feature confusion, and local aggregation. As shown in the sub-images around the original image, large amount of occlusion between objects in the UAV images. And partially occluded objects are unevenly distributed, often gather in a few regions of the UAV image (the regions inside the red boxes of the original image).}
\label{fig1}
\end{figure}

\begin{figure}[t]
\centering
\includegraphics[scale=0.325]{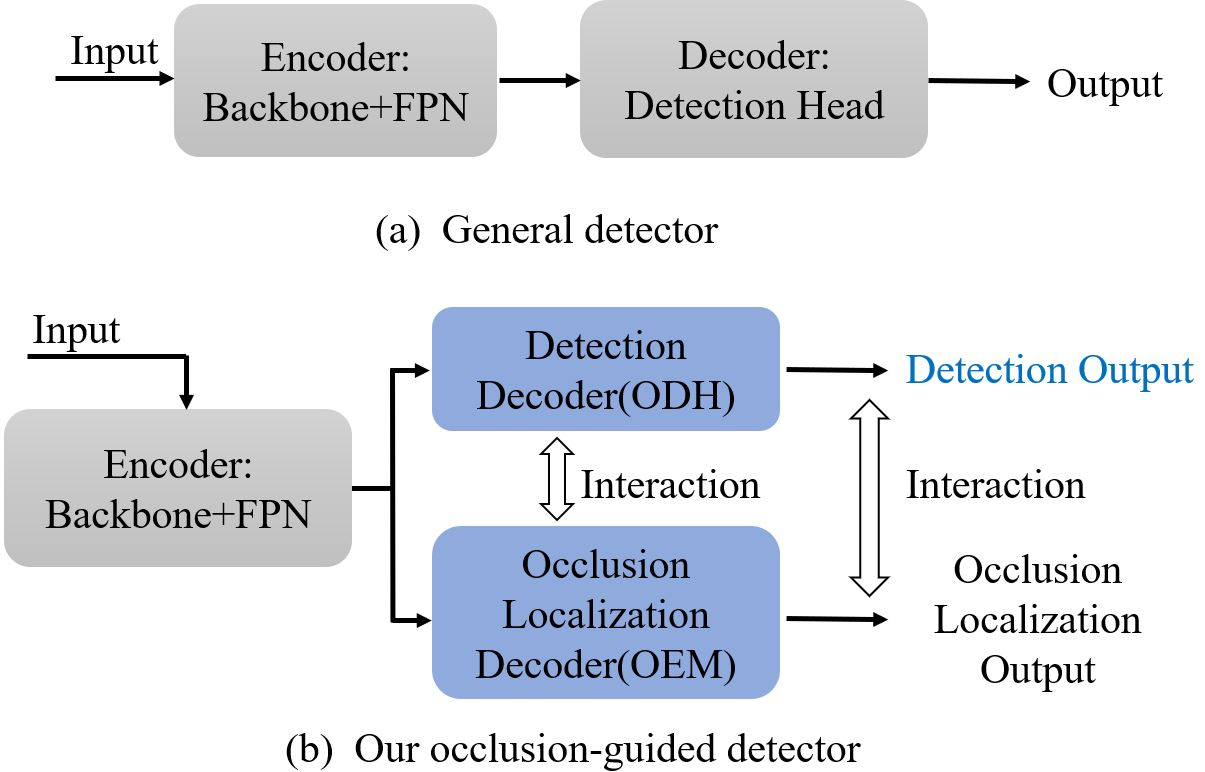}
\caption{General detector and our occlusion-guided detector. A novel occlusion-guided multi-task paradigm is implemented. And we embed a novel occlusion estimation module (OEM) as the occlusion localization decoder and propose a new occlusion decoupling head (ODH) as the detection decoder instead of the original detection head to form our multi-task network.}
\label{fig2}
\end{figure}

The comparison in \autoref{figa} shows that occlusion in UAV scenes such as the Vidrone dataset \citep{du2019visdrone} happens about ten times more than that in natural scenes such as the CityPersons dataset \citep{zhang2017citypersons}. Occlusion with such high-frequency is a great challenge for the detector, but it has been ignored in UAV image object detection. Unlike in natural scenes, the characteristics of occlusion in UAV images can be summarized as follows: (1) \textbf{Local aggregation.} Occlusion is aggregated in UAV images rather than sporadically distributed in natural images \citep{kortylewski2019localizing}. The uneven distribution and local crowding of objects \citep{deng2020global} are issues specific to UAV images. And occlusion happens together with these issues so occlusion also has the characteristic of local aggregation, which is as shown in the red boxed areas of \autoref{fig1}. (2) \textbf{Feature confusion.}  Generic detectors tend to learn discriminative features \citep{zhou2019discriminative}, but discriminative features of occlusion objects are likely to be occluded in spatial location by other occlusion objects. From the camera's shooting viewpoint, the natural scene images are mostly acquired from the side view of the object, and the remote sensing images are mostly acquired from the top view of the object, while the UAV images are in between. Therefore, the object's features in UAV images include both side features of objects and the top features of objects, which are rich and complex features. All these factors make it more difficult for existing detectors to extract semantic feature information of these occlusion areas of occlusion objects. Similar to the feature confusion issue \citep{li2019alleviating}, we name such a problem as feature confusion of occlusion objects. 
Based on these occlusion characteristics, we propose a novel occlusion-guided multi-task network to improve detection performance in UAV images.

As shown in \autoref{fig2}, the general detector predicts detection results from the encoder's output, which lacks perception of occlusion. But addressing the occlusion challenge requires that the detector is able to perceive the occlusion, which is similar to localize occlusion in this image classification work \citep{kortylewski2019localizing}. And the difficulty with this perception is that existing object detection works do not have such an occlusion localization implementation. Therefore, a detector with the ability to localize occlusion is urgently needed. Inspired by the multi-task paradigm \citep{vandenhende2021multi}, we introduce for the first time the occlusion localization task into the detector to form a novel occlusion-guided multi-task network (OGMN) that combines both occlusion localization and object detection. The implementation of the first step gives the model the ability to perceive occlusion, but does not fully exploit it for detection. The introduction of the occlusion localization task enables the model to perceive occlusion, but it is not fully utilized for detection. Therefore, the full and effective use of the occlusion localization results is also an issue. Inspired by multi-task interaction in this work\citep{vandenhende2021multi}, the OGMN designs two new occlusion-guided multi-task interactions to utilize the occlusion localization results in detection. In essence, our OGMN consists of three techniques:

First, to predict occlusion location information, a novel occlusion estimation module (OEM) and corresponding training truth generation method are proposed. As shown in \autoref{fig2}, we embed the OEM as the occlusion localization decoder into the general detector to form a novel multi-tasking network for both the occlusion localization task and the object detection task. Second, we design a multi-task interaction between the two tasks' decoders to address the feature confusion problem of occlusion objects. And we propose a detection decoder occlusion decoupling head (ODH) instead of the original detection head. The ODH decouples occlusion localization results from OEM into encoder output features to generate the decoupling features for the detection task. Finally, to pursue better detection performance, we proposed a two-phase progressive refinement process (TPP) for the detection process to address the local aggregation characteristic of occlusion. The TPP detects an original image and selects several sub-regions where the occlusion objects are aggregated with occlusion localization results. Thus the model refinement detects these occlusion sub-regions and outputs fine detection results, which are merged with the detection results of the original image to form the final detection results. What’s more, the sub-regions selection of TPP is an implicit training data augmentation in the training phase, which makes the model more robust to occlusion objects. Specifically, our proposed ODH and TPP guided by occlusion localization results not only achieve higher accuracy but also improve the robustness of the model to occlusion objects.

Extensive experiments and complete evaluations demonstrate that our proposed approach can solve the poor detection performance caused by the occlusion challenge and is more effective and more robust than the state-of-the-art models. Our approach both achieves better performance on the Visdrones and UAVDT datasets and provides new insights into occlusion in UAV images.
The main contribution of this paper is summarized as follows:

\begin{enumerate}
\itemsep=0pt
\item For the first time, the paper systematically analyzes the occlusion challenge that leads to poor detection performance in UAV image object detection and summarizes it into feature confusion problem and local aggregation characteristic. And we propose an occlusion-guided multi-task network (OGMN) to address the challenge. 
\item To localize occlusion, we introduce the occlusion localization task into the detector and propose a novel occlusion estimation module (OEM) to precisely estimate occlusion location information, and propose an occlusion map truth generation method for the training phase.
\item To fully exploit the occlusion localization results, we present an occlusion decoupling head (ODH) and a two-phase progressive refinement process (TPP) for occlusion-guided multi-task interactions, which achieve higher detection accuracy and more robust detection performance. 
\end{enumerate}

\section{Related Work}
\subsection{Object detection in UAV images}
Object detection in UAV images is essential for traffic monitoring, etc \citep{zhang2021multi,zhang2020identifying,nex2022uav} and is more challenging than natural images. Generic detectors consist of anchor-free detectors \citep{law2018cornernet,duan2019centernet,tian2020fcos} and anchor-based detectors \citep{ge2021yolox,lin2017focal,ren2015faster,cai2018cascade} perform poorly because of scale variability, occlusion and uneven distribution of objects. Feature pyramid structure \citep{lin2017feature} fuses geometric information of shallow features and semantic information of deep features to solve scale variability, and pafpn \citep{liu2018path}, bifpn \citep{tan2020efficientdet} both propose different structures for a better fusion of feature maps of multiple stages. Further, \citep{wang2019spatial} proposes one spatial attention for multi-scale feature refinement for object detection in UAV images. For the inconsistent object distribution challenge, some works \citep{ozge2019power,zhang2019fully,li2020density,yang2019clustered,zhou2019scale,deng2020global} are based on image cropping to solve this challenge. \citep{zhang2019fully} trains a module to estimate the region of small and imbalanced class objects for cropping. Li et al. \citep{li2020density} first estimates the object density map of the UAV image, and then crops it according to the density map. \citep{yang2019clustered} extracts the cluster regions and rescales them for fine detection. \citep{zhou2019scale} and \citep{deng2020global} introduce super-resolution after cropping to improve resolution.
\citep{mittal2022dilated} uses receptive fields and feature maps fusion to propose dilated resnet module. \citep{xiong2022unified} optimizes the training strategies to improve the small-size and long-tail distributed UAV object detection. \citep{kong2022realizing} proposes a location scale
equilibrium to improve the uneven distribution problem. \citep{xi2022fifonet} revitalizes the multi-scale feature representation to improve small-scale problem. 
The transformer-based detectors have also recently gained attention due to global attention benefits. Swin transformer\citep{liu2021swin} introduces the transformer method to object detection. FEA-Swin\citep{museboyina4250755transformer} integrates context information into the swin transformer backbone to form a foreground enhancement attention swin transformer, which is used to UAV object detection.\citep{hendria2021combining} fuses transformer-based and CNN-based models for UAV object detection. The above methods keep exploring for better object detection models but ignore the effects of occlusion and subsequent processing steps for more complex objects contained in cropped regions. This makes the detector only address scale variability and uneven distribution challenge, with occlusion still a challenge. In this paper, the detector crops based on occlusion perception, with the advantage that the guidance of occlusion objects replaces the guidance of the overall objects, which makes the model more robust to the occlusion objects.

\subsection{Occlusion estimation and localization}

Occlusion has a significant impact on computer vision in natural images \citep{wang2018repulsion,kortylewski2019localizing,kortylewski2020compositional}. The reason is that objects surrounded and partially occluded are common. Training loss function and localization of occlusion are two mainstreams to address occlusion challenges. For improving the train loss function, \citep{wang2018repulsion} try to weaken the effect of occlusion by designing new loss functions. They model the occlusion between pedestrians in real-world scenarios with attraction and repulsion of objects. However, the experimental results show that it is satisfactory for pedestrian occlusion, but cannot be adapted to the occlusion of UAV photography objects. Another way of thinking, occlusion localization is significant for solving poor perception of occluded objects. \citep{kortylewski2019localizing,kortylewski2020compositional} design networks to locate occlusion in image classification. Although these works are well located in natural images, they use CompositionalNets. Numerous components of CompositionNets require manual design, and every component in CompositionNets has a great impact on the network. For object detection of UAV images, these natural scene occlusion schemes cannot be migrated to the UAV image detectors because of the complexity of UAV image occlusion.  
Although the occlusion in UAV images which includes Visdrone \citep{du2019visdrone}and UAVDT \citep{du2018unmanned} is more complex than in natural images such as CityPersons \citep{zhang2017citypersons}, this important challenge is overlooked. The paper proposes an occlusion localization network base on a convolutional neural network, that to locate occlusion as an extra decoder. For better network convergence, the paper proposes an occlusion map truth generation method for training. 

\begin{figure*}[t]
\centering
\includegraphics[scale=0.45]{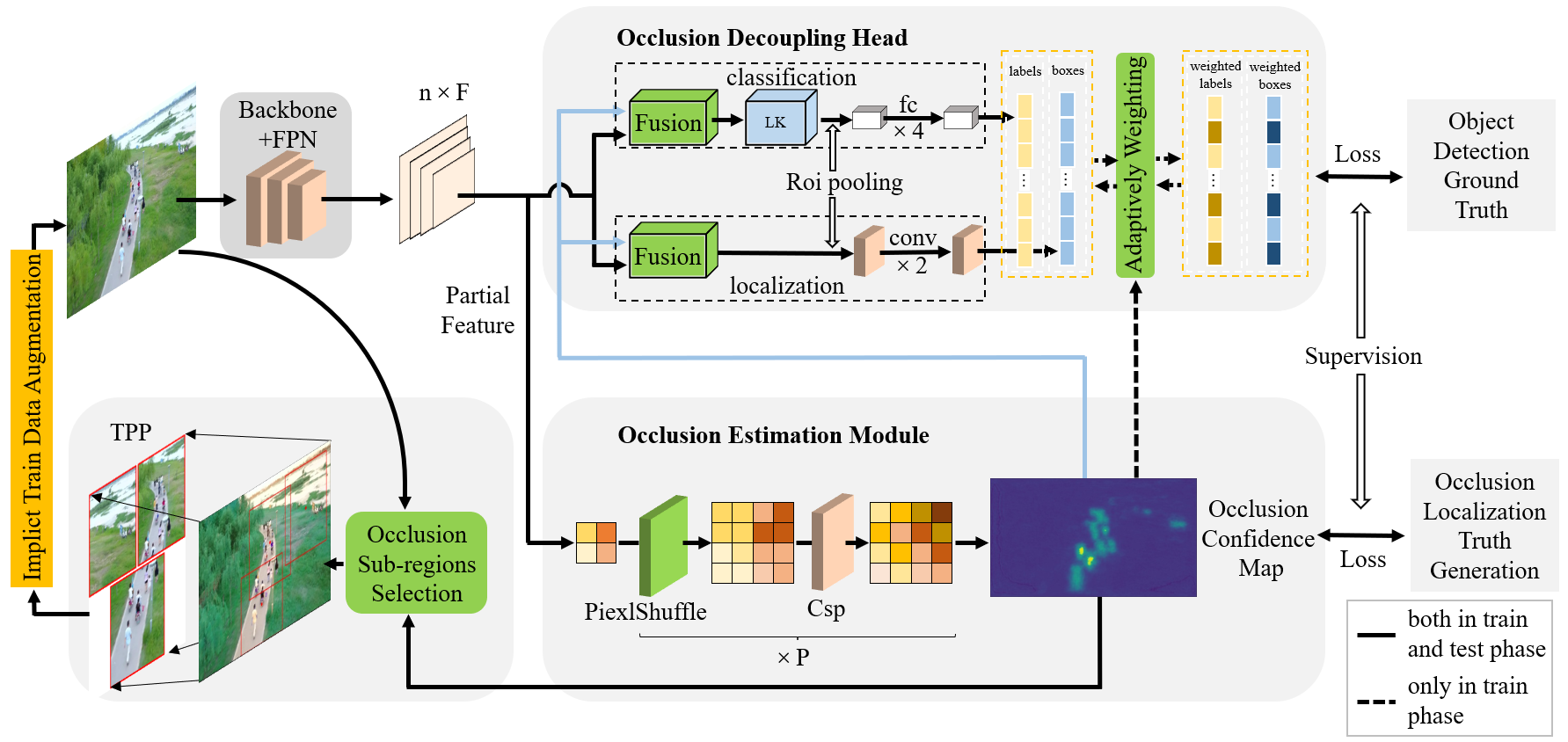}
\caption{The overall structure of our occlusion-guided multi-task network (OGMN). Given an original input image, the OEM predicts an occlusion confidence map of the original image, which is used for the two-phase progressive refinement process (TPP). The TPP selects several occlusion sub-images based on the occlusion sub-regions selection algorithm as the new input images. The ODH implement detection task for original image and sub-images with their occlusion confidence map. Their detection results are merged to form the final detection results of the original input image. The process on the dotted line is only implemented in the training phase. The ‘TPP’ is the two-phase progressive refinement process in our OGMN. The ‘LK’ indicates the large kernel convolution module.}
\label{fig3}
\end{figure*}

\subsection{Decoupling tasks in object detection}
Decoupling classification task and localization task is the core of decoupling tasks for object detection. The reason for decoupling is that there are large biases between the object localization branch and the object classification branch \citep{jiang2018acquisition,song2020revisiting,wu2020rethinking,ge2021yolox}. The vast majority of detectors use sibling heads (shared heads for classification and localization) that combine classification and localization tasks. IoU-Net \citep{jiang2018acquisition} is the first to discover that the sibling's head has problems with classification and localization tasks interfering with each other. Therefore, IoU-Net adds a branch in the head to predict the IoU of the detection bounding boxes and their corresponding ground-truth boxes. \citep{song2020revisiting} observes that the shared head has spatial misalignment for two tasks and classification relies on salient features while bounding boxes localization relies on boundary features, thus they generate two disentangled proposals to decouple classification and localization tasks from spatial dimension. \citep{wu2020rethinking} directly splits the shared head into two branches to achieve decoupling and proposes that convolution layers are more suitable for the localization branch, while fully connected layers are more suitable for the classification branch. \citep{ge2021yolox} proposed YOLOX's decoupling head by combining the methods of the above works. However, the drawback of the above works is that they only decouple the encoder output and do not consider the object feature confusion caused by occlusion. Inspired by this, the paper proposes an occlusion decoupling head, which considers both an inconsistency between classification and localization branches and covers the fusion of occlusion semantic information.

\section{Method}

As shown in \autoref{fig3}, our proposed occlusion-guided multi-task network (OGMN) consists of the following three key techniques: occlusion estimation module (OEM), occlusion decoupled head (ODH) and two-phase progressive refinement process (TPP). The OGMN is designed based on a multi-task paradigm, including occlusion localization and two occlusion-guided multi-task interactions (Sec 3.1).  The OEM localizes occlusion in UAV images and outputs occlusion localization results for occlusion-guided multi-task interactions (Sec 3.2). 
The ODH decouples the occlusion localization results into the classification and localization tasks, and mining occlusion samples by weighting occlusion boxes (Sec 3.3). And the OGMN designs a two-phase progressive refinement process (TPP) to adaptively find out occlusion sub-regions, coarsely detects the original image and finely detects the occlusion sub-images, then merges the coarse and fine detection results with non-maximum suppression (Sec 3.4). The following four subsections in this section detail the proposed techniques.

\subsection{Occlusion-guided Multi-task Network}

To address the occlusion challenge in UAV image object detection to improve detection performance, an occlusion-guided multi-task learning paradigm is adopted to design our occlusion-guided detection network (OGMN), including occlusion localization and two occlusion-guided multi-task interactions. 

The goal of OGMN is to obtain multi-scale features and occlusion spatial location of occlusion objects. Similar to the discussion in \citep{sun2021pbnet}, an occlusion object can be described as a combination of occlusion areas and unocclusion areas. Therefore, an occlusion object with $k$ occlusion areas and $n-k$ unocclusion areas is denoted as:
\begin{equation}
\begin{aligned}
B &=  (l^{occ}_1,\dots,l^{occ}_{k},l^{unocc}_{k+1},\dots,l^{unocc}_{n})  \\
\end{aligned}
\end{equation} 
where $l^{occ}_{i}$ is the $i$-th occlusion area, and $l^{unocc}_{i}$ is the $i$-th unocclusion area. 

The information of occlusion location is described as $F_{occ}$, and the occlusion location depends on $B$. As a result, an occlusion-guided detector to achieve the detection of occlusion objects referring to their occlusion location information. According to the law of Bayes, the detection result can be described as $p(B|F_{occ}, F_{i})$ for multi-scale features $F_{i}$ extracted by the encoder of an input image.
\begin{equation}
\begin{aligned}
p\big(B|F_{occ}, F_{i}\big) &= \frac{p\big(F_{occ}|B,F_{i}\big)\cdot p\big(B|F_{i})}{p(F_{occ}\big)}   \\
\end{aligned}
\end{equation} 
where $p(F_{occ}|B,F_{i})$ is the probability of obtaining occlusion location information given the encoder's feature $F_{i}$ and the detection of occlusion. $p\big(B|F_{i})$ is the probality of obtaining occlusion objects given encoder's feature $F_{i}$.

Assuming the occlusion objects in an input image, the detection result is the position with the greatest posterior probality:
\begin{equation}
\begin{aligned}
B^{\ast} &= \mathop{\mathrm{argmax}}\limits_{B}\Big(p\big(B|F_{occ}, F_{i}\big)\Big)   \\
      &= \mathop{\mathrm{argmax}}\limits_{B}\Big(p\big(F_{occ}|B,F_{i}\big)\cdot p\big(B|F_{i}\big)\Big)
\end{aligned}
\end{equation} 
Therefore, the detection depends on the likelihood model $p(F_{occ}|B,F_{i})$ and the spatial prior $p\big(B|F_{i}\big)$. In this paper, we propose an occlusion-guided multi-task network to model these two factors. Localizing occlusion can model the likelihood model $p(F_{occ}|B, F_{i})$. The spatial prior $p\big(B|F_{i}\big)$ can be modeled as an object detection process.   

Achieving occlusion localization is the core of the OGMN. The strategy of OGMN can be summarized as using the occlusion localization result to guide the detection. The implementation of occlusion localization is a prerequisite for the implementation of this strategy, and an accurate and reliable occlusion localization result is a guarantee for the effectiveness of the whole strategy. But existing object detection works do not have an implementation method for occlusion localization. Therefore, a novel occlusion estimation module is proposed and embedded in the detector to form a new multi-task network that takes into account both the occlusion localization task and object detection task. In the work, new occlusion localization and object detection are two associated tasks in the OGMN, they can share complementary information and act as each other's regularizers, with the potential to improve their respective performance \citep{vandenhende2021multi}. And because the features for the occlusion localization task are partially the same as those required for object detection, they can share one encoder for feature extraction. Thanks to this, the number of parameters in the network and the resulting memory footprints are greatly reduced compared to one encoder per task for two tasks. For the network architecture, the overall network architecture consists of a shared feature extraction encoder and two separate task decoders, which is a multi-task network paradigm as shown in \autoref{fig2}. 

The multi-task interaction is an effective way to improve model performance \citep{vandenhende2021multi}. The decoding phase in our network uses two separate task-specific heads for the two different tasks, which satisfies the independent output of the two tasks, but lacks full interaction. To take full advantage of the occlusion localization results, the ODH and TPP are designed as two occlusion-guided multi-task interactions to utilize the occlusion localization results in the detection to achieve better detection performance. The design of ODH and TPP is based on the characteristics of occlusion in UAV images. To address the feature confusion problem of occlusion objects, we design the ODH instead of the original general detection head. According to the local aggregation characteristic of occlusion, we propose a two-phase progressive refinement process for improving the detection process.

\subsection{Occlusion Estimation Module}
\subsubsection{Occlusion Estimation Module Network}

The occlusion localization task is the core of the OGMN, and accurate occlusion localization results are needed for OGMN to guide the detection. The existing object detection works do not have a method to achieve this requirement. Therefore, a novel occlusion localization network that can adequately express the occlusion localization task is urgently needed. The paper proposes an occlusion localization decoder network that functions as a UAV image occlusion estimator, which is designed by occlusion tiny-scale characteristics of UAV images. According to the purpose, a novel occlusion estimation module (OEM) is designed. 

\begin{figure}[t]
\centering
\includegraphics[scale=0.44]{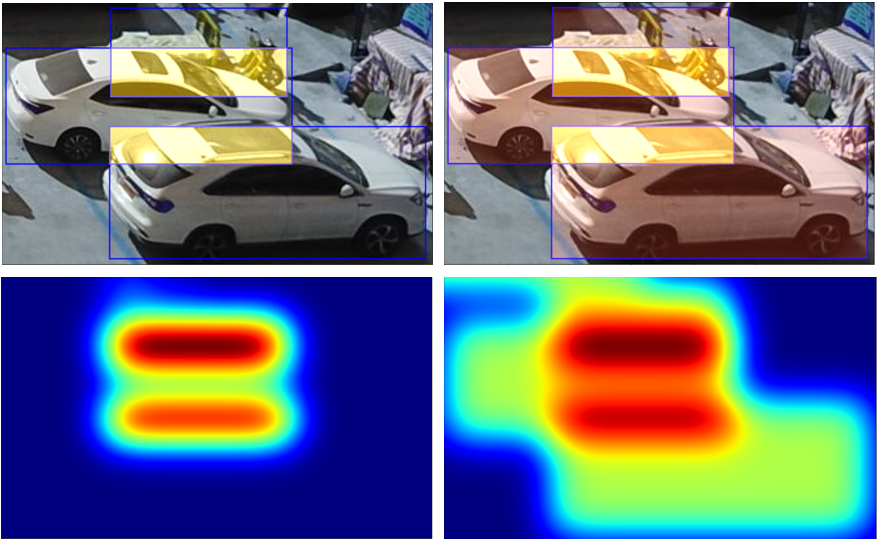}
\caption{The generation of ground truth occlusion maps. The occlusion truth generation based on the ground truth can provide location information of occlusion. Map Only with occlusion (left) has a large gap with the object detection task, resulting in OGMN that is difficult to converge. And the map with highlighting occlusion (right) strengthens the correlation between the occlusion estimation task and the object detection task, which results in a better convergence network.}
\label{fig4}
\end{figure}

The OEM network includes multiple upsampling, which is done to estimate the occlusion more accurately. The reason is that most of the objects in UAV images are small-scale objects, and the occlusion between them is smaller-scale, in addition to the fact that the encoder outputs features after multiple downsampling. To accurately locate occlusion, the OEM upsamples feature map with PixelShuffle \citep{shi2016real} to improve the resolution. The CSP block \citep{ge2021yolox} can interact well with the feature of the current location and the surrounding locations, so the block is used to exploit the local context information after each upsampling. In summary, the alternation of upsampling and convolutional interaction is a design criterion for OEM. \autoref{fig3} shows more details of the OEM network. Partial multi-scale feature maps from the encoder are input to the OEM, and output occlusion confidence maps with occlusion spatial information. Thus each pixel position of the image is given an occlusion confidence value, which constitutes a pixel-level occlusion localization result. 

Consequently, given a multi-scale feature map $F_{i}$ from the encoder in the network, and denote PixelShuffle and CSP in $p$-th series as $U_p(\cdot)$ and $C_p(\cdot)$, the occlusion confidence map $M^{occ}_i$ of OEM output can be formulated as:
\begin{equation}
\begin{aligned}
M^{occ}_i =  \prod_{p=0}^{P}C_{p}\big(U_{p}(F_{i})\big) 
\end{aligned}
\end{equation}
where $P$ denotes the number of modules repeatedly connected in series.

\subsubsection{Occlusion Localization Truth Generation}

After the OEM can adequately express the localization task, another key to determining whether the OEM can output accurate occlusion localization results is the generation of training truth labels. The key to the precise output of a network with sufficient fitting potential is the truth labels of participating network training. According to the principle, the OEM in the network is supervised during the training process. But the available UAV image object detection datasets are not labeled with occlusion locations and do not provide truth labels for OEM training. Therefore, we need to design a new truth label generation method for the supervised training of OEM, and we design a novel generation method of occlusion maps truth labels, that are generated from the object ground truth boxes. 

As shown on the left in \autoref{fig4}, only the overlapping areas of object ground truth boxes are blurred with the Gaussian kernel to make the occlusion map more consistent with the true occlusion distribution. However, after the OEM is trained with such a generation method, the semantic information contained in the output occlusion confidence map is too different from the semantic information required for object detection. And the difference makes it difficult for the network to balance the two tasks in training. This results in poor convergence of the whole network. To better convergence of the network and regress more precise occlusion semantic information, the paper proposes a generation method with highlighting occlusion to generate effective occlusion maps truth labels for training. As shown on the right in \autoref{fig4}, our proposed generation method minimizes the difference by highlighting the occlusion areas within the object ground truth boxes, which strengthens the correlation of the two tasks and achieves better convergence of the network.

The training loss function of the OGMN corresponds to the structure of the multi-task network, which consists of the training loss for two tasks. Given $N$ input images, where the $n$-th input images are $I_{n}$ and the truth occlusion map is $T_{n}$, the loss function for occlusion estimation task $L_{occ}$ is based on mean square error, which is given as follows:
\begin{equation}
\begin{aligned}
L_{occ} &=  \frac{1}{N}\sum_{n=1}^{N}\Big(O\big(E(I_{n})\big) - T_{n}\Big)^2  \\
\end{aligned}
\end{equation}  
where $E(\cdot)$ and $O(\cdot)$ respectively denote the encoder and the occlusion localization decoder in OGMN.

\subsection{Occlusion Decoupling Head}

The design of OEM makes it successful that the occlusion localization task is introduced, and the detection model can perceive occlusion. Although the OEM is able to accurately estimate occlusion localization results, the results are still not directly and fully utilized in the object detection task. Adequate task interaction in the decoder phase is an important way to improve the performance per task \citep{vandenhende2021multi}, while the model still lacks multi-task interactions. And because the introduction of occlusion localization is the first time, the multi-task interaction of object detection and occlusion localization tasks does not exist in the existing work. Therefore, we need to design a new multi-task interaction between two task decoders guided by occlusion localization results. Again, because of the inconsistency between the classification and localization in detection head \citep{wu2020rethinking}, the new multi-task interactions between two task decoders need to design with a decoupling strategy.

In ODH, the proposed decoupling approach independently fuses occlusion localization results to the classification and regression networks. The advantage of decoupling is that the discrepant features required for the two tasks can be extracted. In addition, it is found through experiments that a simple classification network does not perform well. And the ODH introduces the large convolution kernel \citep{ding2022scaling} to increase the receptive field. The network branches of two tasks independently stack image feature map and occlusion feature map in channel dimension, then reduce the number of channels by 1*1 convolution and enhance their respective fused feature with a large kernel convolution module \citep{ding2022scaling}. We denote a multi-scale feature of the encoder output as $F_{i}$, the decoupling approach can be described as the following function. Where $Cat(\cdot)$ and $Conv_{1*1}(\cdot)$ respectively denote concatenation of features, 1*1 convolution with batch normalization layer and ReLU activation. And $LK(\cdot)$ is the large kernel convolution.
\begin{equation}
\begin{cases}
\begin{aligned}
F_{i}^{cls} &= LK\bigg(Conv_{1*1}\Big(Cat\big(F_{i}^{encoder}, O(F_{i})\big)\Big)\bigg)\\
F_{i}^{loc} &= Conv_{1*1}\Big(Cat\big(F_{i}^{encoder}, O(F_{i})\big)\Big)
\end{aligned}
\end{cases}
\end{equation}
where $O(\cdot)$ denotes the occlusion localization decoder in OGMN.

Complementarily, the ODH proposes the sampling strategy of occlusion boxes weighting to mine occlusion hard samples. \citep{lin2017focal,cai2020learning} all focus on hard sample mining, also by weighting the hard sample boxes. But the way they discover hard samples is through training loss. This is not able to select occlusion samples. Therefore, the paper proposes to select occlusion samples from the occlusion confidence map, which regressed from OEM. We denote the occlusion confidence map of the OEM output as $map_{occ}$, which contains the pixel-level occlusion confidence scores. The proposed occlusion hard sample mining is instance-level rather than pixel-level. To convert pixel-level confidence to instance-level confidence, we combine the predicted box coordinates and occlusion confidence map and calculate the sum of the pixel-level confidence scores in the box as the confidence of the instance. And the utilization of occlusion hard sample mining is accompanied by the weighting of the loss function. The classification and localization loss functions for object detection task $L_{cls}$ and $L_{loc}$ is described below:

\begin{equation}
\begin{cases}
\begin{aligned}
L_{cls} &= \frac{1}{N}\sum^{N}_{n=1}L_{n}=- \frac{1}{N}\Big( \sum_{n=1}^{N}\big(\bm{w^{occ}_{n}}\sum_{c=1}^{C}y_{n}^{c}\log{p_{n}^{c}} \big)  \Big) \\
L_{loc} &= \sum_{n}^{N}\Big( \bm{w_{n}^{occ}}\sum_{k\in(x,y,h,w)} smooth_{L_{1}}\big(t_{n}^{k}-v_{n}^{k}\big) \Big)\\
\end{aligned}
\end{cases}
\end{equation}
in which
\begin{equation}
\begin{aligned}
\bm{w^{occ}_{n}} = 
\begin{cases}
2 & {sum(t_{n}, M^{occ}_{n})>=Thr_{occ}}\\
1 & {sum(t_{n}, M^{occ}_{n})< Thr_{occ}}
\end{cases}
\end{aligned}
\end{equation}
where $C$ denotes the number of categories and $y^{c}_{n}$ is a symbolic function that takes the value 1 if the true category of the sample $n$ is equal to $c$ and 0 otherwise; $p^{c}_{n}$ denotes the probability that the sample $n$ belongs to the predicted category $c$. And $v_{n}^{k}\in(t_{n}^{x},t_{n}^{y},t_{n}^{w},t_{n}^{h})$, $t_{n}^{k}\in(t_{n}^{x},t_{n}^{y},t_{n}^{w},t_{n}^{h})$ denote the box coordinates of GT and the predicted box coordinates of sample $n$. $sum(t_{n}, map^{occ}_{n})$ denotes the sum of occlusion confidence scores in $M^{occ}_{n}$ corresponding to the predicted box $t_{n}$ of sample $n$. The $Thr_{occ}$ is the threshold of selection occlusion objects. 

The total training loss function $L_{total}$ is the sum of two tasks with balanced weighting for balancing each task, which is given as follows:
\begin{equation}
\begin{aligned}
L_{total} = \lambda_{occ}L_{occ} + \lambda_{cls}L_{cls} + \lambda_{loc}L_{loc}\\
\end{aligned}
\end{equation}
where $\lambda_{occ}$, $\lambda_{cls}$, and $\lambda_{loc}$ denote the weight parameters of the occlusion localization task, classification task, and localization task in object detection task respectively. $L_{occ}$ is loss function of occlusion localization task in Sec 3.2.2, and $L_{cls}$ and $L_{loc}$ are detection loss functions.

\subsection{Two-phase Progressive Refinement Process}

\begin{figure}[t]
\centering
\includegraphics[scale=0.34]{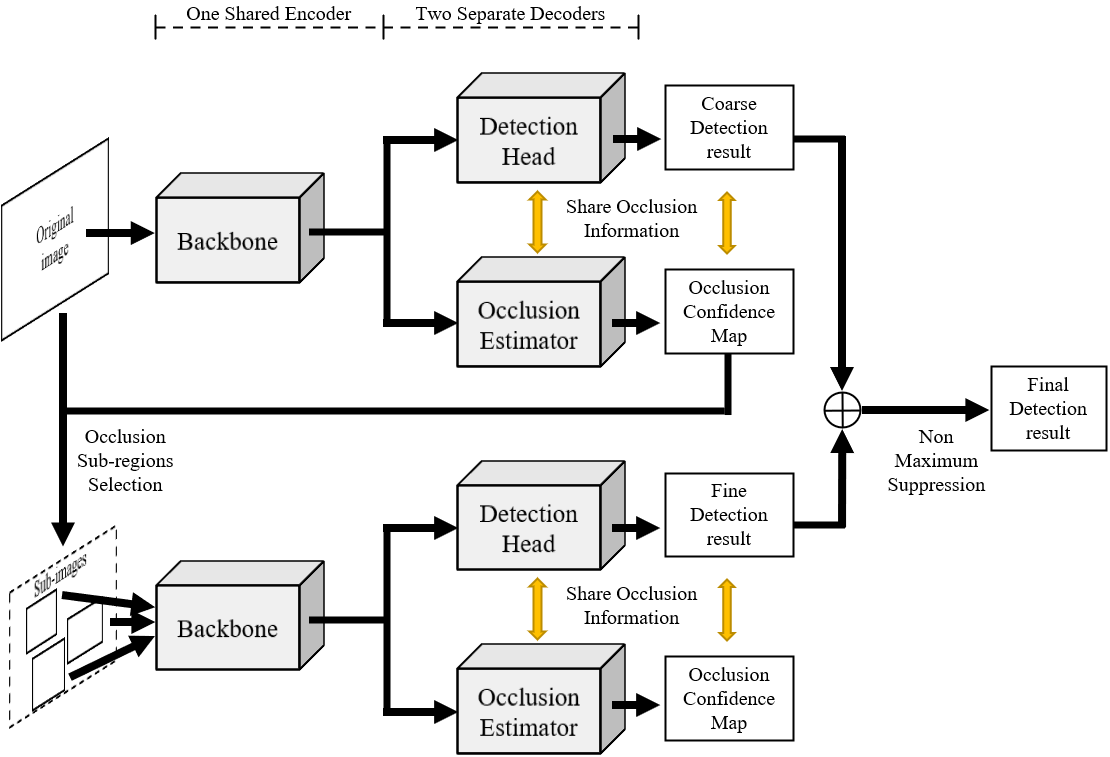}
\caption{Two-phase progressive refinement process in our detector.}
\label{fig5}
\end{figure}

Although the ODH is the occlusion-guided interaction between two task decoders, there is still a lack of the interaction occlusion localization task and detection process. For the detection process in UAV image object detection, the image cropping strategy in these works \citep{li2020density,deng2020global} is another method to improve the detection process, and this strategy is effective to crop out the objects' aggregation regions. But these existing works are overall data-guided, and they ignore the fact that occlusion objects are the hard set for detection. Therefore, an occlusion-guided image cropping between the occlusion localization task and detection process is needed, and we propose a two-phase progressive refinement process (TPP) according to local aggregation characteristic of occlusion as shown in \autoref{fig5}. Compared to existing work \citep{li2020density,deng2020global}, our TPP replaces the guide of the overall data with a guide of occlusion objects, which is similar to the mining hard samples in \citep{shrivastava2016training}. The TPP coarsely detects the original image and finely detects the cropped-out occlusion aggregation regions, constituting a two-phase progressive refinement process for the detection process.

The advantage of TPP is that it is a refine detection process guided by the occlusion objects. In detail, the detection results of OGMN are derived from two phases, including the coarse detection results of the source image and the fine detection results of occlusion sub-regions. The detector first predicts the coarse results of downsampled source image, mainly the bounding box, classes, and confidence of salient objects. After cropping the sub-images corresponding to the occlusion sub-regions selected adaptively by the occlusion confidence map, the detector fine detects these sub-images to output more accurate results. The fine detection phase is the key to improving the detection of hard samples such as occlusion objects. The results of the two phases are merged by NMS to output the final detection results. It is worth mentioning that OGMN is more robust to occlusion objects, 
since the selection of occlusion sub-regions is based on occlusion localization results, and most of the occlusion objects are in these regions, so occlusion objects can be fine detected. This process is extremely helpful to improve the robustness of the model for occlusion objects.

The self-adaptive selection of occlusion sub-regions in the fine detection is documented in Algorithm \autoref{algorithm1}.      
\begin{algorithm}
\caption{Selection of occlusion sub-regions}
\label{algorithm1}
\begin{algorithmic}[1]
\Require{UAV image $I$. Occlusion confidence map $map_{occ}$. Minimum scale of sub\_regions $H_{R},W_{R}$. Window size $H_{w}$ and $W_{w}$. Threshold $Thr$. Number of sub-regions $N$.}
\Ensure{List of occlusion sub-regions coordinates.}\\
\emph{Initialization:}\\
\emph{$Map_{h},Map_{w} \gets I.shape.height, I.shape.weight$}\\
\emph{$Mask_{occ} \gets Zeros(Map_{h},Map_{w})$}\\
\emph{Generate 2D occlusion mask matrix:}
\For {$p$ in $range(0,Map_{h},H_{w})$}
\For {$q$ in $range(0,Map_{w},W_{w})$}
\State $window\_occ \gets Sum(map_{occ}[p:p+H_{w},q:q+W_{w}])$
\If {$window\_occ > Thr$}
\State $Mask_{occ}[p:p+H_{w},q:q+W_{w}] \gets 1$
\EndIf
\EndFor
\EndFor\\
\emph{Output sub\_regions by mask with KMeans clustering:}\\
\emph{$R_{1}...R_{N} \gets Cluster\_by\_mask(Mask\_{occ},N)$}\\
\emph{Correct the coordinates to ensure that all sub-regions are within the original image area, and the scale is higher than the minimum scale value:}\\
\emph{$R_{1}...R_{N} \gets Correct\_sub\_regions([R_{1}...R_{N}],(H_{R},W_{R}))$}\\
\Return $[R_{1}...R_{N}]$
\end{algorithmic}
\end{algorithm}

\begin{table*}[width=1.0\textwidth,cols=4,pos=b]
\scriptsize
\renewcommand{\arraystretch}{1.5}
\caption{The objects density and overlap density among the UAV images dataset and natural occlusion images dataset. The threshold of overlap is IOU greater than 0.5.}
\label{table1}
\centering
\begin{tabular}{c|c|c|cll}
\toprule
\cline{1-4}
Scene           &Dataset                                & objects per image & overlaps per image    \\ \cline{1-4}
\multirow{2}{*}{UAV images}     &Visdrone\citep{du2019visdrone}         & 73.30             & 3.93        \\ 
                &UAVDT\citep{du2018unmanned}            & 14.83             & 1.02           \\
\midrule
\midrule
Natural images  &CityPersons\citep{zhang2017citypersons}& 6.47              & 0.32        \\
\bottomrule
\end{tabular}
\end{table*} 
We denote an original input image as $I$  and the final detection result is $R_{final}$, the detection result is approximated by:
\begin{equation}
\begin{aligned}
R_{final} = NMS[D(E(I),O(E(I))),\\ \bigcup_{q=0}^{Q}D(E(I_{q}), O(E(I_{q})))] \\ 
\end{aligned}
\end{equation}
where $E(\cdot)$ denotes the encoder in the network for extracting multi-scale feature maps, and $D(\cdot)$ and $O(\cdot)$ respectively denote the object detection decoder and the occlusion localization decoder, which is detailed are Sec 3.3 and Sec 3.2. 
$I_{0},\cdots,I_{N}$ denote occlusion sub-images, which are selected with algorithm 1: $I_{0},\cdots,I_{Q} = Select(D(E(I)), I)$.    

The size of the sub-regions is determined by the scale difference of the objects in the images. The statistics of the datasets show that the larger objects in the UAV images are more than four times the scale of the smaller objects. Therefore, in order to eliminate as much as possible the scale differences of objects in the image, the minimum value of the size of the sub-regions is determined as one-quarter of the image size.

At the same time, in order to ensure that the divided sub-images can cover most of the occlusion objects, the maximum value of the sub-regions size is not fixed. As described in Algorithm \autoref{algorithm1}, the sub-regions generated using the k-means clustering algorithm are already guaranteed to cover the perceived occlusion objects, and on the basis of these sub-regions coordinates, the size of the sub-regions is corrected according to the minimum value constraint. This ensures that the sub-regions cover the perceived occlusion objects, while eliminating scale differences between objects in the images.

\begin{figure}[t]
\centering
\includegraphics[scale=0.417]{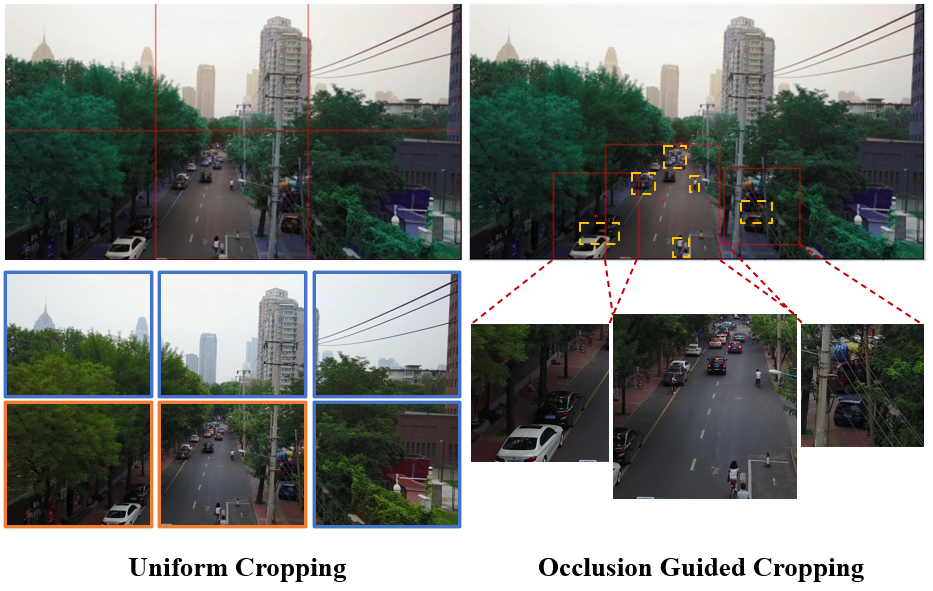}
\caption{The comparison of uniform cropping (left) and occlusion-guided cropping (right). (1) Uniform cropping: most of the cropped sub-region (Blue rectangle) do not contain objects, and only a few discriminative sub-regions (Orange rectangle); (2) Occlusion guided cropping: crop few sub-regions based on occlusion position (Rectangular position of yellow dotted line), containing the majority of objects. In addition, implicit training data augmentation for hard samples including occlusion objects is implemented.}
\label{fig6}
\end{figure}

Implicit training data augmentation is incorporated in our proposed two-phase progressive refinement process. The fine detection results of occlusion sub-regions are merged with the detection results of the original image in the testing phase, while in the training phase are used to optimize the model. This implicit training data augmentation is similar to generic data augmentation techniques such as uniform cropping and random cropping. But these generic techniques are overall data-guided, which lack the perception of occlusion. While our method perceives the occlusion well and guides the model to focus on the occlusion objects in the training phase by implicit training data augmentation. This makes the model more robust for occlusion hard samples. The implicit training data augmentation in our TPP can be described as follows: 

The occlusion sub-regions are selected by Algorithm \autoref{algorithm1} with occlusion localization results. 
During the training phase, these regions are cropped out and resized to a suitable scale as new training data $\bigcup_{q=1}^{Q}I_{q}$, and together with the source images $I$ to form the whole training data $I\bigcup(\bigcup_{q=1}^{Q}I_{q})$.

\section{Experiments and evaluations}

In this section, the paper evaluates the proposed techniques from different perspectives. Intuitively, an optimized detection algorithm designed for a subnet of occlusion samples not only performs for this specific subnet of samples but also significantly improves the whole data. Our evaluation focuses on two aspects: the whole dataset and the occlusion objects contained therein. 

\subsection{Datasets}

Robustness to arbitrary instance distributions is a requirement for an ideal detector, i.e. not only effective for clustered occlusion objects but also stable to detect sparse occlusion objects. The paper adopts two datasets: Visdrone \citep{du2019visdrone} and UAVDT \citep{du2018unmanned} for the complete evaluation of heavily and slightly overlapped scenes concurrently.  The \autoref{table1} lists the “instance density” of the two datasets in the experiments and CityPersons dataset \citep{zhang2017citypersons}. The Citypersons dataset is widely used for the study of occlusion in natural scenes. And the comparison shows that occlusion happens more frequently in UAV images than in natural scenes. Since our proposed approach in the paper mainly aims to address the occlusion challenge in UAV images. Therefore, the experiments perform most of the ablations on the Visdrone, which has heavily overlapped instances. Note that the experiments on UAVDT are to verify whether the proposed approach is effective to detect slightly overlapped objects.

Visdrone.The Visdrone2019-DET dataset \citep{du2019visdrone} contains images acquired by the UAV at different angles and different heights, consisting of 6471 training images, 548 validation images, and 3190 testing images. The resolution of the image is distributed between 1920*1080 and 960*540. And the dataset provides bounding boxes of instance and ten predefined fine-grained categories labels (pedestrian, car, bicycle, et al.) for training images and validation images manually annotated. Occlusion instances are a large part of these annotation instances. Taking the validation set as an example, the percentage of instances with occlusion areas over 0.4, 0.3, and 0.2 are 10.9\%, 19.2\%, and 29.6\% respectively. And such a large number of occlusion instances is the basis of our proposed occlusion-aware detection. Therefore, the experiments fully evaluate the effectiveness of our proposed approach on the dataset. Although the occlusion ratio is provided in the dataset annotation, there is no information on the location of occlusion. So the paper only uses the occlusion ratio annotation for the evaluation of occlusion instances, not for model training. Since the annotations of the testing set are not provided, our approach is trained on the training set and evaluated on the validation set, and the related works \citep{zhang2019fully,li2020density,yang2019clustered,deng2020global,liu2021hrdnet} are always done in this way.

UAVDT. The UAVDT dataset \citep{du2018unmanned} includes over 40000 aerial images from UAVs acquired at varying heights and angles, and their resolutions are 1024*540 and 960*540. The instances in the images are labeled with three categories(car, truck, and bus) and bounding boxes. The occlusion labeling provided by the dataset indicates that the percentage of occlusion instances is 11.4\%, and these occlusion instances are sufficient to support the valid evaluation of our approach. Although the percentage is not higher than the Visdrone dataset, we only conduct the comparative evaluation with other advanced works on the dataset. 

\subsection{Evaluation Metrics}
The paper mainly takes two criteria for different purposes:

Similar to the most relevant object detection works \citep{xu2022detecting,wei2020oriented,weber2021artificial}, $AP$ (mean Averaged Precision), $AP_{75}$ (APs at the IoU threshold of 0.75) and $AP_{50}$ (APs at the IoU threshold of 0.5) are used to evaluate our approach on the overall datasets, which are the most popular evaluation metric for object detection. Since AP is a comprehensive indicator of precision and recall, the AP value calculated for the whole data reflects the detection performance for the overall dataset, including both occlusion instances and normal instances. They are the most used metrics in other UAV image object detection works. Larger AP and $AP_{50}$ and $AP_{75}$ indicates better performance.

Averaged Recall (AR) \citep{everingham2010pascal} is used to evaluate the occlusion samples in the ablation experiment, which is the recall ratio corresponding to a fixed number of boxes in the image. If calculating AR only for the occlusion instances of one image, the harm of false detection results from other instances can be avoided. Thus when calculating AR only for occlusion instances in the ground truth, such evaluation metric accurately reflects whether the detector can recall more occlusion instances, which is denoted as $AR_{occ}$. Larger $AR_{occ}$ indicates better performance. 

In addition, the APs and AR for different scales of evaluation metrics \{$AP_{s}$, $AP_{m}$, $AP_{l}$, $AR_{s}$, $AR_{m}$, $AR_{l}$ \} are used in ablation experiments. 

\begin{table*}[width=1.0\textwidth,cols=4,pos=t]

\scriptsize
\renewcommand{\arraystretch}{1.5}
\centering
\caption{Accuracy comparison with the state-of-the-art models on the validation set of the Visdrone dataset. Our models only trained on the Visdrone train set. And we do not use any other training tricks or testing tricks. ‘--’ indicates that the result is not reported by the corresponding work. ‘RC’ indicates random cropping. $\star$ indicates that we have not found a suitable name for the model. ‘ResNeXt101+50’ is ResNeXt101 network and ResNet50 network.}
\label{table2}
\begin{tabular}{c|c|c|c|c}
\toprule
\cline{1-5}
\multirow{2}{*}{Method}       &\multirow{2}{*}{Backbone} &\multirow{2}{*}{AP} &\multirow{2}{*}{$AP_{50}$} &\multirow{2}{*}{$AP_{75}$}        \\ 
 & & & & \\                                   
\cline{1-5} 
\midrule
Cascade RCNN\citep{cai2018cascade}(CVPR)            &ResNeXt101        &29.7    &51.6       &30.3      \\ \cline{1-2}
Cascade RCNN+RC\citep{chen2019mmdetection}          &ResNeXt101        &29.8    &51.9       &29.9      \\ \cline{1-2}
$\star$\citep{zhang2019fully}(ICCV)              &ResNeXt152        &30.3    &--         &--        \\ \cline{1-2}
ClusDet\citep{yang2019clustered}(ICCV)        &ResNeXt101        &28.4    &53.2       &26.4      \\ \cline{1-2}
DMNet\citep{li2020density}(CVPR)          &ResNeXt101        &29.4    &49.3       &30.6      \\ \cline{1-2}
GLSAN\citep{deng2020global}(TIP)           &ResNet101         &32.5    &55.8       &33.0      \\ \cline{1-2}
HRDNet\cite{liu2021hrdnet}(ICME)         &ResNeXt101+50     &33.5    &56.3       &34.0      \\ \cline{1-2}
AdNet\citep{zhang2021multi}(ISPRS)	         &Darknet-53        &31.1    &57.9       &30.5      \\
\midrule
\midrule 
OGMN(Ours)                          &ResNet101         &33.4    &57.7       &33.5      \\ \cline{1-2} 
OGMN(Ours)                          &ResNeXt101        &\textbf{35.0}    &\textbf{59.7}       &\textbf{35.8}      \\ \cline{1-2}
\cline{1-5}
\bottomrule
\end{tabular}
\end{table*}

\begin{table*}[width=1.0\textwidth,cols=4,pos=t]

\footnotesize
\renewcommand{\arraystretch}{1.5}
\centering
\caption{Accuracy comparison with the state-of-the-art models on the test set of the UAVDT dataset. And we do not use any other training tricks or testing tricks. ‘--’ indicates that the result is not reported by the corresponding work. $\star$ indicates that we have not found a suitable name for the model.}
\label{table3}
\begin{tabular}{c|c|c|c|c}
\toprule
\cline{1-5}
\multirow{2}{*}{Method}       &\multirow{2}{*}{Backbone} &\multirow{2}{*}{AP} &\multirow{2}{*}{$AP_{50}$} &\multirow{2}{*}{$AP_{75}$}        \\ 
 & & & & \\                              
\cline{1-5} 
\midrule
$\star$\citep{zhang2019fully}(ICCV)              &ResNet50         &15.1     &--         &--          \\ \cline{1-2}
ClusDet\citep{yang2019clustered}(ICCV)        &ResNet50         &13.7     &26.5       &12.5        \\ \cline{1-2}
DMNet\citep{li2020density}(CVPR)          &ResNet50         &14.7     &24.6       &16.3        \\ \cline{1-2}
GLSAN\citep{deng2020global}(TIP)           &ResNet50         &19.0     &30.5       &21.7        \\ \cline{1-2}
CDMNet\citep{duan2021coarse}(ICCV)         &ResNet50         &16.8     &29.1       &18.5        \\ \cline{1-2}
ARMNet\citep{wei2020amrnet}			  &ResNet50         &18.2     &30.4       &19.8        \\ \cline{1-2}
AdNet\citep{zhang2021multi}(ISPRS)	      &Darknet-53       &21.3     &\textbf{42.6}       &18.3        \\ 
\midrule
\midrule 
OGMN(Ours)                          &ResNet50         &20.9     &34.5       &23.2        \\ \cline{1-2}
OGMN(Ours)                          &ResNeXt101       &\textbf{24.2}     &39.9       &\textbf{26.8}        \\ \cline{1-2}
\cline{1-5}
\bottomrule
\end{tabular}
\end{table*}

\subsection{Implement Details}

Baseline model: The experiments are to verify the proposed approach in the MMDetection framework \citep{chen2019mmdetection}. Cascade RCNN \citep{cai2018cascade} is used as the baseline detector with the pre-trained weightings on ImageNet \citep{deng2009imagenet}. Because our approach has instance-level occlusion object selection, the selection algorithm can only be based on the proposals of the two-stage detection models, so we choose the popular two-stage detection model Cascade RCNN as our baseline model. On the Visdrone dataset, the backbone of Cascade RCNN is ResNet101 \citep{he2016deep} and ResNext101 \citep{xie2017aggregated} respectively. The model with ResNet101 is used to compare with other state-of-the-art models with the same backbone, while the model with ResNeXt101 is used for ablation studies and compared with other state-of-the-art models with the same backbone. On the UAVDT dataset, the Cascade RCNN is implemented with both ResNeXt101 and ResNet50 \citep{he2016deep}. When the backbone is ResNet50, The results of models are mainly compared with other state-of-the-art models, while the results of the model with ResNeXt101 backbone show the better detection ability for our method. 

Our OGMN: Our OGMN is a multi-task detection network based on the baseline model. The whole network consists of a feature extraction encoder and two novel decoders: the occlusion decoupling head (ODH) for object detection task, and the occlusion estimation module (OEM). Based on the whole multi-task network, a two-phase progressive refinement process (TPP) is constructed by several algorithms such as the selection of occlusion sub-regions, which can find out occlusion aggregation regions and refine detect occlusion hard objects in these regions.     

Train phase: For the training of the Visdrone and UAVDT datasets, we used the multi-scale training. The input image size is set to \{1360*765, 1024*765, 1024*1024\} both in the Visdrone dataset and UAVDT dataset in the multi-scale training. The TPP in our proposed approach contains the implicit training data augmentation. For hyperparameters in the TPP, the number of sub-images is preset to 3 in Visdrone and UAVDT datasets, and the detail is discussed in Sec 4.6.3. As a result, the total numbers of training images in the Visdrone and UAVDT datasets are increased to 25368 and 33908, respectively.  Our OGMN and baseline model is both trained on 2 GPUs with the SGD optimizer for 16 epochs in total. The learning rate is fixed at 0.05. 

Test phase: For the test and inference on Visdrone and UAVDT datasets, we use the single-scale, and the size of input images is preset to 1024*1024. And the threshold and detection number per image of NMS are preset to 0.5 and 500, respectively. For hyperparameters in our TPP, the number of sub-images is preset to 3 in line with the training phase. 

Both in training and inferring phases, the parameters of the selection algorithm of occlusion sub-regions \{$H_{R}$, $W_{R}$, $H_{w}$, $W_{w}$, $Thr$\} are preset to \{300, 300, 40, 40, 45\}, respectively. The $P$ in function 4 is 2. The $Thr_{occ}$ in function 8 is 45. The $\lambda_{occ}$, $\lambda_{cls}$ and $\lambda_{loc}$ in function 9 are \{1.0, 1.0, 0.5\}, respectively.

\subsection{Comparative evaluation with state-of-the-arts}

\begin{table}[width=1.0\textwidth,cols=4,pos=t]

\footnotesize

\caption{Accuracy comparison with the latest models and transformer-\newline based models on the Visdrone and UAVDT dataset. And we do\newline not use any other training tricks or testing tricks. ‘--’ indicates\newline that the result is not reported by the corresponding work.}
\centering
\renewcommand{\arraystretch}{1.5}
\label{table4}
\begin{tabular}{c|c|c|c|c}
\toprule
\cline{1-5}
\multirow{2}{*}{Method}       &\multicolumn{2}{c|}{Visdrone}  &\multicolumn{2}{c}{UAVDT}        \\  \cline{2-5}
 &AP &$AP_{50}$   &AP    &$AP_{50}$  \\                              
\cline{1-5} 
\midrule
Swin\citep{liu2021swin}        &29.1        &--     &--       &--    \\ \cline{1-1}
Swin-RFP\citep{hendria2021combining}        &34.4        &59.0     &--       &--    \\ \cline{1-1}
FEA-Swin\citep{museboyina4250755transformer}         &33.7         &--     &--          &--  \\ \cline{1-1}
$\Delta$\citep{xiong2022unified}           &37.8        &61.7     &21.2         &37.4   \\ \cline{1-1}
DCRFF\citep{mittal2022dilated}       &35.0       &57.0    &--     &--        \\ \cline{1-1}
LSEM\citep{kong2022realizing}			  &28.0        &44.9    &--  &--              \\ \cline{1-1}
FiFoNet\citep{xi2022fifonet}	      &36.9      &63.8     &21.3    &36.8         \\ 
\midrule
\midrule 
OGMN(Ours)                          &35.0       &59.7     &24.2    &39.9          \\ \cline{1-1}
\cline{1-5}
\bottomrule
\end{tabular}
\end{table}

Visdrone. In \autoref{table2}, we conduct a quantitative comparison between other state-of-the-art models and our model on the Visdrone dataset validation set. For \citep{zhang2019fully} and HRDNet \citep{liu2021hrdnet}, although the number of parameters and FLOPS of our backbone ResNeXt101 is much lower than their backbone (ResNeXt152\citep{xie2017aggregated}, ResNeXt101 with ResNet50), our detection performance is better than theirs. Specifically, ClusDet, DMNet, and GLSAN use a detection strategy of coarse-fine combination among original images and sub-images. Compared with GLSAN, our model’s detection performance is 0.9\% AP higher when the backbone is the same as ResNet101. And compared with ClusDet and DMNet, our model significantly outperforms them with the same backbone ResNext101, with 6.6\% improvement and 5.6\% improvement.   

\begin{figure*}[b]
\centering
\includegraphics[scale=0.146]{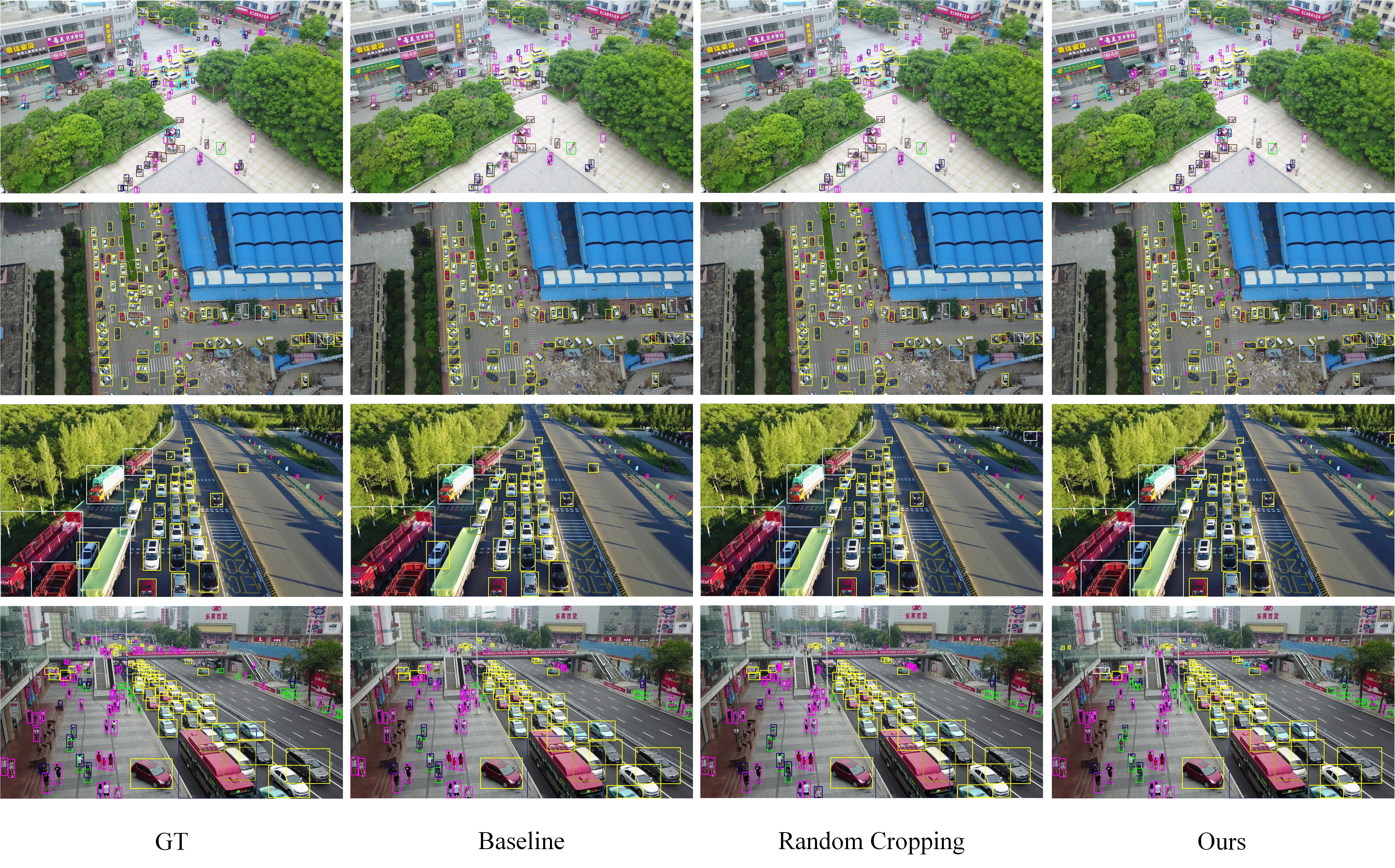}
\caption{The visualization of detection results among ground truth, the baseline of Cascade RCNN, random cropping, and our model on the Visdrone dataset validation set.  }
\label{fig7}
\end{figure*}

UAVDT. In \autoref{table3}, the accuracy comparison result with other state-of-the-art models on the test set of the UAVDT dataset is shown. The evaluation of our OGMN shows a satisfactory improvement. Compared with other state-of-the-art models with the same backbone ResNet50, our OGMN achieves a new state-of-the-art performance: AP is 20.9\%, $AP_{50}$ is 34.5\% and $AP_{75}$ is 23.2\%. And the accuracy improvement compared to previous state-of-the-art models GLSAN is AP: 2.9\%, $AP_{50}$: 4.0\%, and $AP_{75}$: 1.5\%. When the backbone is massive ResNext101, our OGMN achieves better accuracy performance: AP is 24.2\%, $AP_{50}$ is 39.9\% and $AP_{75}$ is 26.8\%. And the accuracy improvement compared to previous state-of-the-art models GLSAN is AP: 5.2\%, $AP_{50}$: 9.4\% and $AP_{75}$: 5.1\%.    

In \autoref{table4}, we compare our model and several transformer-based detectors and other latest models on the Visdrone dataset or UAVDT dataset. Compared to transformer-based detectors such as Swin Transformer\citep{liu2021swin}, Swin-RFP\citep{hendria2021combining} and FEA-Swin\citep{museboyina4250755transformer}, our proposed OGMN performs significantly better than these transformer-based methods. The accuracy of our OGMN implementation of AP 35.0\% is higher than the accuracy of these three methods: 29.1\%, 34.3\%, and 33.7\%. This shows that our proposed OGMN is superior to transformer-based detectors.      

Compared to other latest models, the AP 35.0\% of our OGMN is higher than AP 28.0\% of LSEM\citep{kong2022realizing}. Our OGMN has the same AP 35.0\% as DCRFF\citep{mittal2022dilated}, but achieves a higher $AP_{50}$ (59.7 higher than 57.0). For \citep{xiong2022unified} and FiFoNet\citep{xi2022fifonet}, while our method performs slightly worse on the Visdrone dataset, our method performs significantly better than them on the UAVDT dataset (24.2\% higher than 21.2\% and 21.3\%). The UAVDT dataset is more crowded and smaller scale than the Visdrone dataset, which suggests that our OGMN is more robust than crowded scenes and small-scale object detection. The above analyses show that our OGMN demonstrates competitive detection performance.

\subsection{Visualization analysis}

The qualitative comparison of detection visualization results among the baseline detector(Cascade RCNN), random cropping, and our proposed approach is shown in the \autoref{fig7}. It is observed that our approach outperforms other detection models in all fine-grained categories, especially in categories with more occlusion, such as people, and bicycle. The outstanding performance is mainly due to the detection network guided by occlusion estimation results. In addition, our approach can accurately detect a portion of occlusion instances that are not annotated in the ground truth, such as occlusion pedestrian in \autoref{fig7} first column. This suggests that our model is a more robust detector for occlusion objects, although it does harm to calculating AP. What’s more, it can be observed that our model performs well for different scales and different classes, and is also robust to both scales and classes. 

\subsection{Ablation studies}
\begin{figure*}[t]
\centering
\includegraphics[scale=0.53]{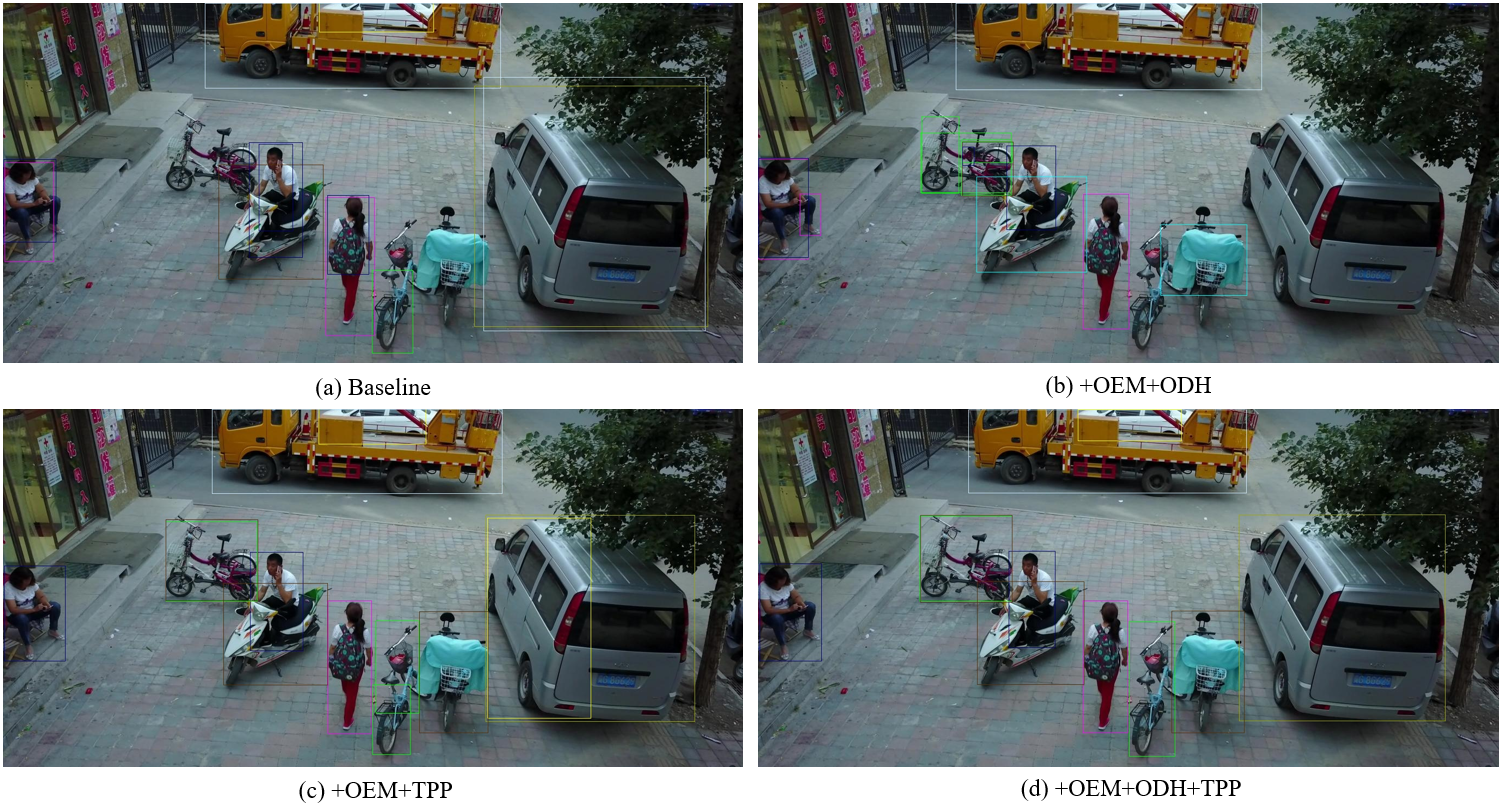}
\caption{The visualization of our ablation detection results compared to the baseline when adding different components. The experiments are conducted on the Visdrone dataset validation set.}
\label{fig8}
\end{figure*}

\begin{figure}[h]
\centering
\includegraphics[scale=0.56]{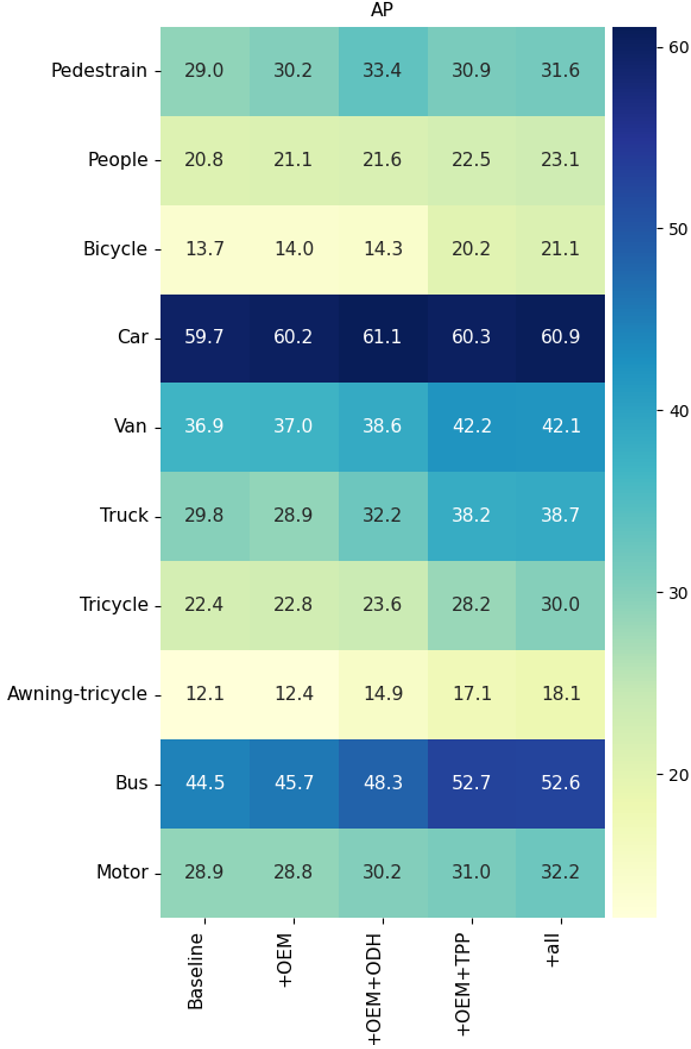}
\caption{The ablation studied of ten fine-grained categories with AP on the Visdrone validation set. We add OEM, ODH, and TPP components based on the baseline model Cascade RCNN. The implementation of ODH and TPP is dependent on OEM. ‘all’ indicates that all three components are added. The darker the color, the higher the accuracy.}
\label{fig9}
\end{figure}

To verify the contribution of each component in our proposed approach, this subsection conducts thorough ablation experiments on the validation set of the Visdrone dataset. And the baseline is fixed at Cascade RCNN with the ResNeXt101 backbone. The accuracy results of the ablation experiments in Table IV are for the whole instances in the dataset, while the visualization results of ablation experiments are shown in \autoref{fig8}. And the accuracy results of the ablation experiments for the fine-grained categories are shown in \autoref{fig9}.

\begin{table*}[width=1.0\textwidth,cols=4,pos=t]

\footnotesize
\renewcommand{\arraystretch}{1.5}
\centering
\caption{Accuracy comparison among the baseline models of Cascade RCNN, and models with the different components in our approach on the validation set of Visdrone dataset. The implementation of ODH and TPP is dependent on OEM.}
\label{table5}
\begin{tabular}{c|c|c|c|c|c|c|c|c|c|c|c}
\toprule
\cline{1-12}

\multirow{2}{*}{OEM}  &\multirow{2}{*}{ODH}  &\multirow{2}{*}{TPP} &\multirow{2}{*}{$AP$} &\multirow{2}{*}{$AP_{50}$} &\multirow{2}{*}{$AP_{75}$}&\multirow{2}{*}{$AP_{s}$}&\multirow{2}{*}{$AP_{m}$}&\multirow{2}{*}{$AP_{l}$}&\multirow{2}{*}{$AR_{s}$}&\multirow{2}{*}{$AR_{m}$}&\multirow{2}{*}{$AR_{l}$}      \\
                                        & & &   & & & &  & &  &  &   \\ 
\cline{1-12} \midrule
          &         &         &29.7         &51.6          &30.3          &21.1          &39.9          &45.5          &37.9           &56.7          &65.1     \\
 $\surd$  &         &         &30.2         &52.4          &30.6          &22.6          &38.8          &39.4          &39.8           &55.3          &57.3     \\ 
 $\surd$  & $\surd$ &   	  &31.7         &54.2          &32.4          &23.8          &41.2          &46.3          &\textbf{42.9}  &58.8          &63.2   \\
 $\surd$  &         & $\surd$ &34.4         &59.1          &34.6          &25.7          &47.5          &58.2          &38.9           &60.6          &69.2   \\ 
 $\surd$  & $\surd$ & $\surd$ &\textbf{35.0}&\textbf{59.7} &\textbf{35.8} &\textbf{26.8} &\textbf{47.7} &\textbf{59.4} &40.1           &\textbf{61.9} &\textbf{71.0} \\
\cline{1-12}
\bottomrule
\end{tabular}
\end{table*}

\autoref{fig8} shows the visualization detection results of the ablation experiments by adding our components based on the baseline. For the visualization detection results of baseline in \autoref{fig8} (a), both missing alarm detections and false alarm detections heavily exist, which seriously do harm to APs evaluation metrics. This is because the baseline model lacks the perception of occlusion between objects, and the feature confusion characteristic causes the feature extraction is not sufficient. So the baseline model misses the real occlusion objects and wrongly detects the false occlusion objects. In \autoref{fig8} (b) with occlusion estimation module (OEM) and occlusion decoupling head (ODH), several occlusion objects are recalled and misclassifications are corrected. They are due to the feature supplementation and enhancement capabilities of ODH by adding precise occlusion location feature information from OEM. And in \autoref{fig8} (c) with occlusion estimation module (OEM) and two-phase progressive refinement process (TPP), very few missing alarm detection objects are recalled, but false alarm detections are still not eliminated. This is because the TPP is beneficial to improve the model robustness for occlusion objects by selecting sub-regions where occlusion objects are aggregated, but the occlusion feature information is not used in the detection process. The problem of feature confusion still exists, and the model still lacks perception of occlusion. It can be observed in \autoref{fig8} (d) with all components including OEM, ODH, and TPP, recall for missing alarm detections and false alarm elimination both perform perfectly. The feature supplement by fusion of OEM output occlusion feature in ODH compensates for the deficiency caused by feature confusion, and the selection of occlusion sub-regions in TPP improves model robustness for occlusion objects. The combination of all components results in the advantages of each being exploited by the model.    
 
\begin{figure}[t]
\centering
\includegraphics[scale=0.72]{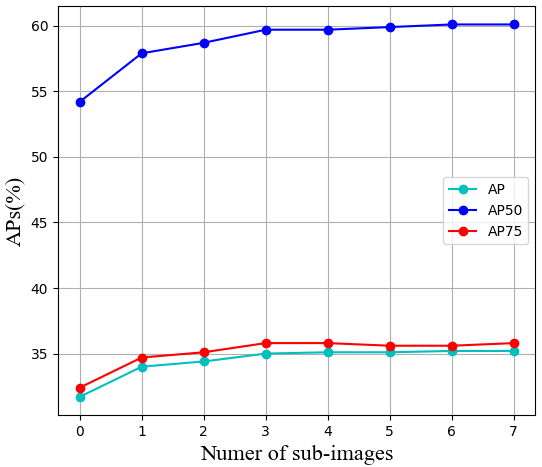}
\caption{The AP, $AP_{50}$ and $AP_{75}$ over the different numbers of sub-images, which is a hyperparameter in our proposed two-phase progressive refinement process. The experiments are conducted on the Visdrone dataset validation set.}
\label{fig10}
\end{figure}

\subsubsection{\textbf{Effect of Occlusion Estimation Module (OEM)}}
In order to implement our occlusion-aware detection approach, the first step is to obtain information about the location of the occlusion between objects in the UAV image. So we add a task decoder OEM for estimating the occlusion location results based on the baseline detection network to form our multi-task network. And the multi-task network is the basis for the implementation of other components in our approach. According to \autoref{table5}, the testing of our multi-task network only with OEM improves AP from 29.7\% to 30.2\%, $AP_{50}$ from 51.6\% to 52.4\% compared to baseline detection networks, which only adds a decoder branch for occlusion location task. In contrast to the baseline with only the detection task, our multi-task model focuses on the performance of two tasks simultaneously. The improvement in AP metrics indicates that our multi-task network with OEM can balance the two tasks well. In detail, our multi-task model completes the occlusion location task without compromising the performance of the object detection task, and in addition, the addition of the occlusion location task branch results in an 0.5\% AP improvement in the performance of the object detection task. The reason for this improvement is that added occlusion localization task by OEM has a strong correlation with the object detection task. Although the decoders of the two tasks do not yet interact, the two tasks contribute to each other during the end-to-end training process, resulting in an initial enrichment of the features needed for each. In summary, our multi-task model by adding OEM and the proposed occlusion location labels generation method are effective.

\subsubsection{\textbf{Effect of Occlusion Decoupling Detection Head (ODH)}}Our occlusion decoupling detection head (ODH) is implemented based on the multi-task network with OEM. Leveraging the estimation results of occlusion localization in the object detection process is a multi-tasking interaction to improve detection performance. Our ODH is designed on this idea. Based on the OEM-based multi-tasking network, \autoref{table5} incorporates the ODH for the ablation experiment. In contrast to the baseline model and our multi-tasking network without ODH, the multi-tasking network with ODH perform better in two aspects. 1) Compared with our multi-tasking network without ODH, the model with ODH improves $AP$ from 30.2\% to 31.7\% and $AP_{50}$ from 52.4\% to 54.2\%, which indicates our ODH is able to improve the ability of the model to find out objects. \autoref{fig9} shows that our ODH is able to improve AP in all ten categories, especially the four categories pedestrian, Truck, awning-tricycle, and bus are improved by 2.4\%, 2.3\%, 2.5\%, and 2.6\%. This indicates that our ODH has great adaptability to all categories. 2) Compared with our multi-tasking network without ODH, our multi-tasking model with ODH improves $AP_{s}$ from 22.6\% to 23.8\%, and the improvement is 1.2\%; $AP_{m}$ and $AP_{l}$ contributes more, which are respectively 2.4\% and 6.9\%. It indicates that our ODH is robust to objects at different scales. This shows that the supplement of features can improve the model's perception of the objects, and the ODH decoupled incorporation of OEM predicted occlusion semantic features is able to solve the feature confusion problem.  

In summary, our proposed ODH is effective for the UAV image object detection task and robust to different classes and scales.

\begin{table*}[width=1.0\textwidth,cols=4,pos=b]
\footnotesize
\renewcommand{\arraystretch}{1.5}
\centering
\caption{The $AR_{occ}$ metric calculation is only for different occlusion objects set of our model on the Visdrone dataset validation set. ‘Accuracy$\nearrow$’ indicates the accuracy improvement compared to the baseline model of Cascade RCNN. }
\label{table6}
\begin{tabular}{c|c|c|c|c|c|c|c}
\toprule
\cline{1-8}
\multirow{2}{*}{metric}    &\multirow{2}{*}{Occlusion Ratio} &{Baseline} &\multicolumn{2}{c|}{Baseline+OEM+ODH}   &\multicolumn{2}{c|}{Baseline+OEM+ODH+Two-phase}  &GLSAN\\ \cline{3-8}                     
                           &                                 &Accuracy   &Accuracy           &Accuracy$\nearrow$ &Accuracy &Accuracy$\nearrow$ &Accuracy \\
\midrule
\midrule
\multirow{3}{*}{$AR_{occ}$}&No Occlusion(0.0)               &86.71\%         &90.83\%&($\nearrow$ 4.12\%)   &\textbf{91.49\%} &($\nearrow$ 4.78\%) & 88.59\%  \\ \cline{2-8}
                           &Partial Occlusion(0.01-0.5)     &79.05\%         &83.68\%&($\nearrow$ 4.63\%)   &\textbf{83.97\%} &($\nearrow$ 4.92\%) & 80.12\%\\ \cline{2-8}
                           &Heavy Occlusion(0.5-1.0)   	 &68.37\%         &73.26\%&($\nearrow$ 4.89\%)   &\textbf{74.92\%} &($\nearrow$ 6.55\%) & 69.03\%\\ 
\cline{1-8}
\bottomrule
\end{tabular}
\end{table*}

\begin{figure}[t]
\centering
\includegraphics[scale=0.73]{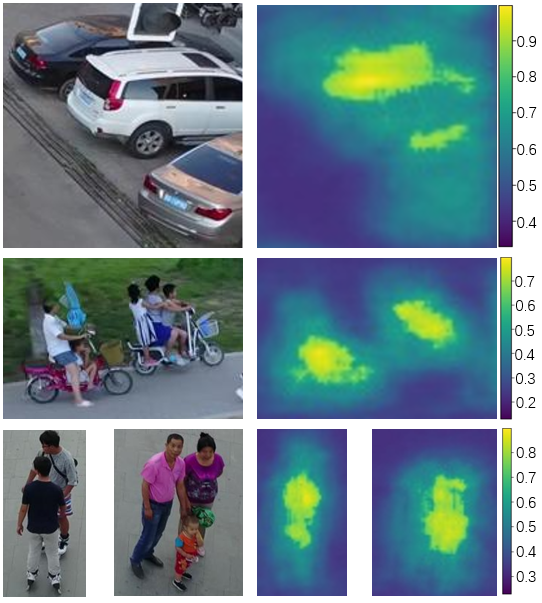}
\caption{The occlusion instances between objects in Visdrone dataset images are shown on the left side. The images on the right side are occlusion confidence maps precisely predicted by the occlusion localization decoder, and the highlighting areas are occlusion location areas.}
\label{fig11}
\end{figure}

\subsubsection{\textbf{Effect of Occlusion Guided Two-phase progressive refinement process (TPP)}}
The large number of occlusion objects in the UAV images and their local aggregation characteristics motivation for occlusion-guided cropping, and thus a two-phase progressive refinement process (TPP) based on occlusion-aware including implicit training data augmentation is designed.
The implementation of our TPP relied on the multi-task network with OEM, because the selection sub-regions selection algorithm of the TPP depended on the occlusion confidence map of the OEM prediction. It can be observed from \autoref{table5} and \autoref{fig9} that the model with TPP and OEM performs better on APs metrics both in the whole data and the fine-grained categories. 1) For the whole data, the multi-tasking network with TPP improves $AP$ from 30.2\% to 34.4\%, $AP_{50}$ from 52.4\% to 59.1\% and $AP_{75}$ from 30.6\% to 34.6\% compare to the model only with OEM, and the APs improvements achieve 4.2\%, 6.7\% and 4.0\% on $AP$, $AP_{50}$ and $AP_{75}$. The great improvement indicates our TPP is able to improve model robustness for the whole data by selecting occlusion regions with occlusion confidence maps predicted from OEM. 2) the model with TPP compared to the model only with OEM, the AP improvement is achieved on all ten fine-grained categories, in particular, 9.3\% $AP$ improvement for truck class, 7.2\% $AP$ improvement for bus class, and 6.2\% $AP$ improvement for bicycle class. 
This great improvement in fine-grained categories indicates our proposed TPP is beneficial to improve the model robust in fine-grained categories. In summary, our proposed TPP is effective for the robustness improvement of the model both in the whole dataset and fine-grained categories.

The number of sub-images is a hyperparameter in the two-phase progressive refinement process, which determines the number of occlusion sub-regions selected by the TPP both in the training phase and inference phase. \autoref{fig10} shows the effect of this hyperparameter preset value on the model detection performance. As it increases from 0 to 3, $AP$, $AP_{50}$ and $AP_{75}$ increase more; after greater than 3, $AP$ and $AP_{75}$ have a small increase and $AP_{75}$ remains stable. A larger hyperparameter improves the model detection performance better, while a smaller hyperparameter makes the model inference faster. To better balance the two aspects, the hyperparameter of the number of sub-images is finally determined as 3 in the training and inferring phases.

\subsection{Analysis for addressing occlusion challenge}
The paper addresses the occlusion challenge to solve the poor detection limitation caused by occlusion in UAV image object detection.
 This subsection analyzes the effectiveness of our proposed approach to addressing the occlusion challenge in three aspects: occlusion localization task, qualitative analysis of occlusion objects, and quantitative analysis of occlusion objects.

\subsubsection{Qualitative analysis for occlusion localization task}

The occlusion localization task is central to the implementation of our proposed methods. Our model improves detection performance in the detection process by the guidance of occlusion localization task results.
 The quality of the output results of the occlusion localization task determines the degree of detection performance improvement. Therefore, the qualitative occlusion location results of our OGMN are shown in \autoref{fig11}. The existing UAV datasets without annotation for occlusion location make quantitative evaluation impossible to perform, the qualitative method is utilized to verify the effectiveness of our model. 
We show the real occlusions on the left side of \autoref{fig11} and the corresponding occlusion location results including occlusion confidence maps and scores on the right side. It can be observed from the highlighting areas in \autoref{fig11} that our approach is able to precisely locate the occlusion between instances such as scar with the car, people with people and people and motor, etc. Accurate prediction and display of output results improve the explainability of the detection process. In summary, our proposed OEM (Occlusion Estimation Module) is able to accurately predict the occlusion location information, and the occlusion location labels generation method and the corresponding loss function are able to satisfy that our multi-task network with OEM converges well during the training process; in short, our proposed approach is effective for the occlusion location task, and the output the occlusion estimation decoder is accurate.

\subsubsection{Quantitative analysis for object detection of occlusion instances}

In \autoref{table5}, our OGMN improves the AP metric over the baseline model (Cascade RCNN) by 5.6\%, but it only shows that our OGMN is effective for the overall data, which contains both unocclusion objects and occlusion objects. Although the average precision in different iou thresholds and several other metrics such as $AR_{s}$, $AR_{m}$ and $AR_{l}$ are the most used evaluation metrics for object detection evaluation, these metrics are designed for the evaluation of the overall dataset, so they can not indicate that if our proposed approach is effective for occlusion objects in the whole UAV dataset. The reason for this is that the predicted unocclusion objects in detection results are harmful to this AP evaluation metric, and these unocclusion objects are relatively false alarm detection results for predicted occlusion objects. Therefore, if we want to evaluate the effectiveness of the detector for the occlusion objects, it is necessary to calculate a separate metric only for occlusion objects set in the overall dataset.
To evaluate the effectiveness of our proposed approach for the occlusion objects, the novel metric $AR_{occ}$ for occlusion instances is proposed.

\begin{figure}[t]
\centering
\includegraphics[scale=0.95]{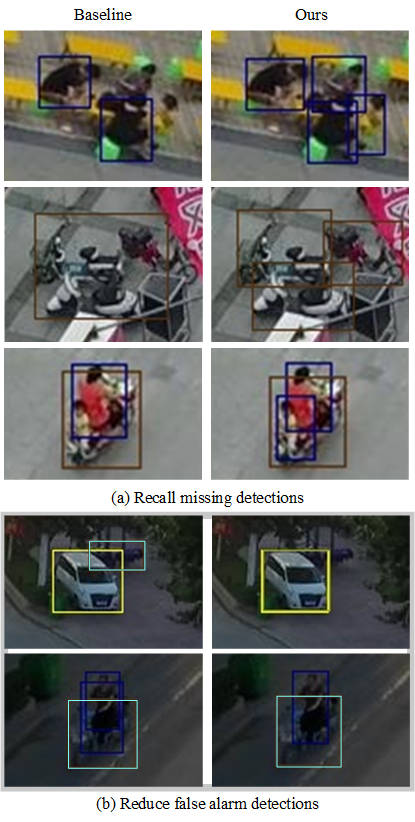}
\caption{The visualization of our model’s detection results compared to the baseline on occlusion instances. Recalling more missing detections and reducing false alarm detections are two major advantages of our model.}
\label{fig12}
\end{figure}

Cleverly, we adopt the recall rate only for occlusion instances $AR_{occ}$ as the evaluation to satisfy the requirements: do not harm by predicted unocclusion detection results and be able to show the effectiveness of the detector on occlusion instances. The evaluation results show in \autoref{table6}. The occlusion ratios are the corresponding annotation provided in the Visdrone dataset, which according to the occlusion ratio divides the annotated ground truth box into three parts. When the occlusion ratio of occlusion objects are 0.0\%, they are no occlusion object, while from 1\% to 50\% are partial occlusion objects, and from 50\% to 100\% are heavy occlusion objects. The $AR_{occ}$  metric calculates the average recall for each of these three parts of data with different occlusion ratios. It can be observed that the $AR_{occ}$ metric improvement by our proposed approach is mainly in two aspects: the improvement compared to the baseline model and the detection performance improvement of occlusion objects. 

Compared to the baseline, our OGMN with all three components has greatly improved the recall of all three parts of the data with different degrees of occlusion: the $AR_{occ}$ improvement for no occlusion data is 4.78\%, the $AR_{occ}$ improvement for partial occlusion data is 4.92\% and the $AR_{occ}$ improvement for heavy occlusion data is 6.55\%. Such improvements show that our proposed approach is effective for all data with different degrees of occlusion including unocclusion objects and occlusion objects, and both can improve the model's perception of them.
From the comparison of the $AR_{occ}$ improvement of the three parts data with different degrees of occlusion, the 6.55\% $AR_{occ}$ improvement for heavy occlusion objects is higher than the improvement for partial occlusion objects and no occlusion objects, and the 4.92\% $AR_{occ}$ improvement is higher than the improvement for no occlusion objects. This indicates that our proposed approach improves the detection performance by addressing the occlusion challenge, and the detection of the occlusion data is improved better than unocclusion. 

In addition, we evaluate the inference results of the GLSAN method with this metric and add the evaluation results to \autoref{table6} for comparison with our method. The comparison results show that although GLSAN improves compared to the baseline, the improvement is mainly greater for the subset of lightly occlusion objects, which is more evidence that our method is more robust to occlusion objects.

In short, our OGMN is indeed proposed for the occlusion challenge and is more robust to the occlusion data.    

\begin{figure*}[t]
\centering
\includegraphics[scale=0.26]{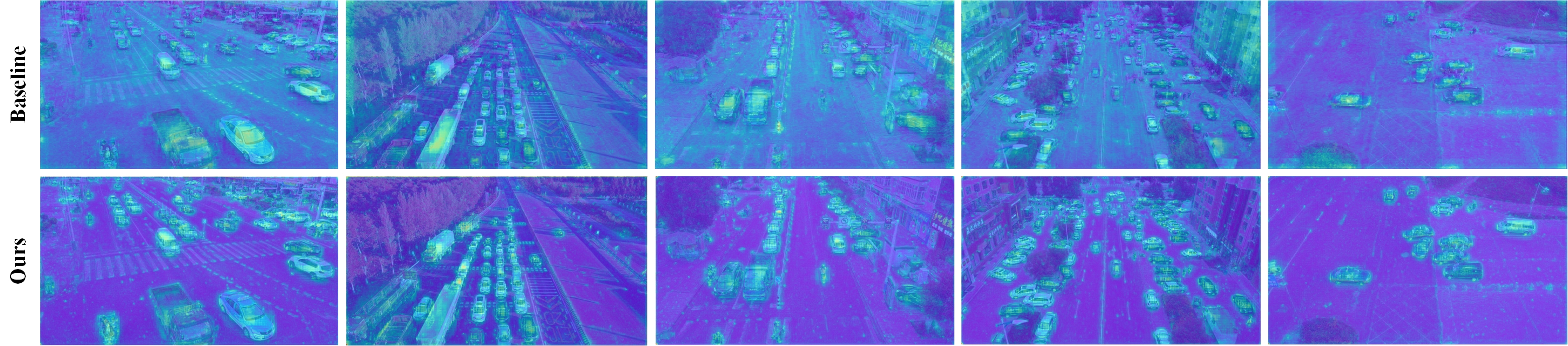}
\caption{The visualization of feature maps among baseline and our model on the Visdrone dataset validation set. Compared to the baseline model, our model perceives more discriminative areas of the objects and is able to suppress interference from background areas well.}
\label{fig13}
\end{figure*}

\subsubsection{Qualitative analysis for object detection of occlusion instances}

The visualization of the detection results for occlusion objects in the Visdrone validation set is shown in \autoref{fig12}. It can be observed that our approach contributes mainly to recalling more occlusion objects and reducing the false alarm detection of apocryphal occlusion objects. For the former, recalling more real occlusion objects facilitates the improvement of the precision rate; and for the latter, reducing false alarm detection objects in the baseline model is beneficial for improving the precision rate. Because the two metrics' precision rate and recall rate directly determine the $AP$ metrics, our proposed approach can achieve a great improvement on the $AP$ metrics value. In summary, our approach is effective and robust for occlusion objects compared to the baseline model.

\begin{figure}[t]
\centering
\includegraphics[scale=0.46]{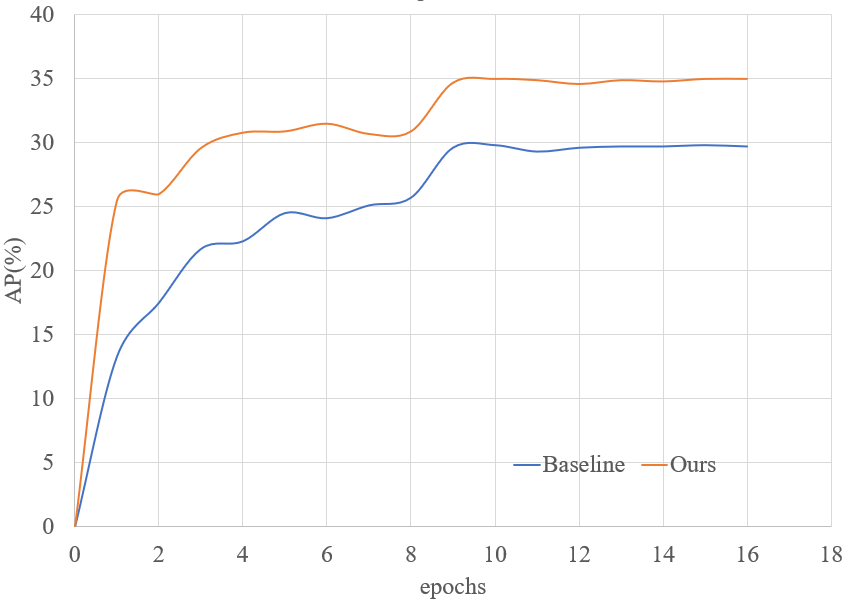}
\caption{AP training convergence curves for our model and the baseline model on the Visdrone dataset validation set.}
\label{fig14}
\end{figure}

\subsection{Feature Map Analysis}

As shown in \autoref{fig13}, we conduct the visualization of feature maps among baseline and our model. The baseline model focuses on extracting a small number of discriminative areas of objects in UAV images, and incorrectly extracts some background areas as discriminative features. This is the reason why the performance of existing detectors is limited.
Thanks to the guidance of the occlusion estimation task, our model perceives more discriminative areas of the objects instead of a few discriminative areas and is able to suppress interference from background areas well compared to the baseline model.
 Thus our model is able to recall more objects and reduce false alarm detections, achieving higher accuracy and more robust detection performance.

\begin{table}[width=1.0\textwidth,cols=4,pos=t]
\footnotesize
\renewcommand{\arraystretch}{1.5}
\centering
\caption{Comparison of model space complexity with other models on\newline the Visdrone dataset.  }
\label{table7}
\begin{tabular}{c|c|c|c}
\toprule
\cline{1-4}
 Method &AP &Params(M) &Flops(GFLOPs) \\                              
\cline{1-4} 
\midrule
Baseline        &29.7       &137.4     &121.9       \\ \cline{1-1}
+OEM            &30.2       &139.7     &131.3       \\ \cline{1-1}
+OEM+ODH        &31.7       &141.8     &133.2       \\ \cline{1-1}
+OEM+TPP        &34.4       &139.7     &143.6       \\ \cline{1-1}
+OEM+ODH+TPP    &35.0       &141.8     &147.9       \\
\cline{1-4}
\bottomrule
\end{tabular}
\end{table}

\begin{table}[width=1.0\textwidth,cols=4,pos=t]
\footnotesize
\renewcommand{\arraystretch}{1.5}
\centering
\caption{Comparison of inference time with other models on the\newline UAVDT dataset.}
\label{table8}
\begin{tabular}{c|c|c|c}
\toprule
\cline{1-4}
 Method  &GLSAN &FiFoNet &OGMN(ours) \\                              
\cline{1-4} 
AP          &19.0      &21.3       &24.2  \\
\cline{1-4}
Time(s)     &0.76      &0.51       &0.61        \\ 
\bottomrule
\end{tabular}
\end{table}

\subsection{Space Complexity and Time Complexity Analysis}

Space complexity analysis. We conduct the analysis of our OGMN space complexity in \autoref{table7} on the UAVDT dataset. The Params and FLOPs of our proposed OGMN are 141.8 M and 147.9 GFLOPs, which is an increase of only 4.4 M and 26.0 GFLOPs compared to the baseline model 137.4 M and 121.9 GFLOPs. According to the statistics on the Params, the increase of our OGMN space complexity is only 3.2\%. Although our increase in the number of parameters is only 3.2\%, our increase in accuracy is AP 5.3\%. This scale of spatial complexity is sufficient to support the deployment of our proposed OGMN on relevant UAV devices. 

Time complexity analysis. We conduct the analysis of our OGMN time complexity in \autoref{table8} on the UAVDT dataset. Our model and other models for comparison are tested for computational time complexity in the PyTorch environment. GLSAN and FiFoNet have inference speeds of 0.76 s/image and 0.51 s/image respectively, while our OGMN has inference speeds of 0.61 s/image. According to the comparison results, the speed performance of our proposed OGMN is within the same level as that of existing methods. But in terms of accuracy, our OGMN is significantly better than other models. On the UAVDT dataset, the accuracies of GLSAN and FiFoNet are Ap: 19.0\% and AP: 21.3\% respectively, but our OGMN achieves AP: 24.2\%, outperforming the other models.  For actual deployments, a model with this level of inference speed is capable of meeting the requirements of real-time applications in actual deployments after processing such as acceleration by the TensorRT tools.

\subsection{Model Convergence Analysis}  

\autoref{fig14} shows the AP training convergence curves comparison of our model and baseline. From the final convergence results, our model converges to a much higher AP than the baseline, which is the most important, implying that the parameters of the network are better adapted to the detection task. In addition, our model jumps out of the local minimum region better. The baseline falls into the local optimum region at the 4th epoch, 6th epoch, and 8th epoch, and although it escapes from these local optimum regions afterward, it does not contribute to the overall convergence of the model. Fortunately, our model only falls into the local optimum region twice and achieves better convergence afterward. According to the above analysis, our model and approach have better convergence compared to the baseline.  

\section{Conclusion and future work}
In the paper, we proposed an occlusion-guided network for better object detection performance in UAV images. We designed a multi-task network to accomplish occlusion localization task and object detection task. 
Besides, we develop the two-phase progressive refinement detection process and occlusion decoupling detection head, which are guided by the occlusion estimation results based on our designed multi-task network. 
The effective occlusion localization gives our approach the potential to mine out occlusion hard samples for training data augmentation toward further performance in a two-phase process. The evaluations on two datasets (Visdrone and UAVDT) demonstrate the effectiveness, robustness, and characteristics of our proposed approach.

In the future, we will try to address the limitation of Monomodal input data to expect better detection performance in UAV images.
The proposed techniques mainly focus on the occlusion problem in the UAV image detection task.
Despite the excellent performance above, the proposed approach still has one modal data.
Currently, UAV image object detection works rely singularly on RGB images, lacking the assistance of other modal data.
By introducing other modal data such as radar and infrared, the model will be the potential to capture information such as depth, which is advantageous for addressing scale, occlusion, and other challenges in UAV image object detection. In addition, the introduction of multi-modal data facilitates the transformation of implicit modeling to explicit modeling for challenges, which is more intuitive and easy to visualize.

\bibliographystyle{cas-model2-names}

\bibliography{cas-refs.bib}

\end{sloppypar}
\end{document}